\documentclass[sn-basic]{sn-jnl}

\usepackage{graphicx}
\usepackage{multirow}%
\usepackage{amsmath,amssymb,amsfonts}%
\usepackage{amsthm}%
\usepackage{mathrsfs}%
\usepackage[title]{appendix}%
\usepackage{xcolor}%
\usepackage{textcomp}%
\usepackage{manyfoot}%
\usepackage{booktabs}%
\usepackage{algorithm}%
\usepackage{algorithmicx}%
\usepackage{algpseudocode}%
\usepackage{listings}%
\usepackage{makecell}
\usepackage{verbatim}
\usepackage{stfloats}
\usepackage{textcomp}
\usepackage{array}

\newcommand{\Eref}[1]{Eq.~{\ref{#1}}}
\newcommand{\Sref}[1]{Sec.~{\ref{#1}}}

\newcommand{\Tref}[1]{Tab.~{\ref{#1}}}
\newcommand{\Fref}[1]{Fig.~{\ref{#1}}}

\usepackage{xcolor}

\begin{document}

\title[Article Title]{RehearsalNeRF: Decoupling Intrinsic Neural Fields of
Dynamic Illuminations for Scene Editing}


\author[1]{\fnm{Changyeon} \sur{Won}}\email{cywon1997@gm.gist.ac.kr}
\equalcont{These authors contributed equally to this work.}

\author[1]{\fnm{Hyunjun} \sur{Jung}}\email{hyunjun.jung@gm.gist.ac.kr}
\equalcont{These authors contributed equally to this work.}

\author[1,3]{\fnm{Jungu} \sur{Cho}}\email{jungu.cho@cj.net}

\author[1]{\fnm{Seonmi} \sur{Park}}\email{bluesky1000@gm.gist.ac.kr}

\author[3]{\fnm{Chi-Hoon} \sur{Lee}}\email{chi.lee@cj.net}

\author*[2]{\fnm{Hae-Gon} \sur{Jeon}}\email{earboll@yonsei.ac.kr}

\affil[1]{\orgdiv{Department of AI Convergence}, \orgname{GIST}, \orgaddress{ \city{Gwangju}, \country{Korea}}}

\affil[2]{\orgdiv{Department of Artificial Intelligence}, \orgname{Yonsei University}, \orgaddress{ \city{Seoul}, \country{Korea}}}

\affil[3]{\orgdiv{AI R\&D Division}, \orgname{CJ Corporation}, \orgaddress{\city{Seoul}, \country{Korea}}}


\abstract{Although there has been significant progress in neural radiance fields, an issue on dynamic illumination changes still remains unsolved. Different from relevant works that parameterize time-variant/-invariant components in scenes, subjects' radiance is highly entangled with their own emitted radiance and lighting colors in spatio-temporal domain. In this paper, we present a new effective method to learn disentangled neural fields under the severe illumination changes, named \textit{RehearsalNeRF}. Our key idea is to leverage scenes captured under stable lighting like rehearsal stages, easily taken before dynamic illumination occurs, to enforce geometric consistency between the different lighting conditions. In particular, RehearsalNeRF employs a learnable vector for lighting effects which represents illumination colors in a temporal dimension and is used to disentangle projected light colors from scene radiance. Furthermore, our RehearsalNeRF is also able to reconstruct the neural fields of dynamic objects by simply adopting off-the-shelf interactive masks. To decouple the dynamic objects, we propose a new regularization leveraging optical flow, which provides coarse supervision for the color disentanglement. We demonstrate the effectiveness of RehearsalNeRF by showing robust performances on novel view synthesis and scene editing under dynamic illumination conditions. Our source code and video datasets will be publicly available. Project Page: \url{https://wcy199705.github.io/RehearsalNeRF}}





\maketitle

\section{Introduction}\label{sec1}

    Neural Radiance Fields (NeRFs)~\citep{mildenhall2021nerf} represent a scene as neural implicit functions and enable to render photo-realistic images from arbitrary viewpoints.
    For wider applicability of NeRFs, their variant representations for dynamic motions~\citep{Wu2022d2nerf, zhang2023detachable, park2021nerfies, park2021hypernerf, li2021neural, pumarola2021d, weng2022humannerf, peng2021neural} have been actively studied. 
    Existing dynamic radiance fields synthesize sequential frames with novel viewpoints by decoupling static and dynamic objects~\citep{Wu2022d2nerf, zhang2023detachable}, topological deformation~\citep{park2021nerfies, park2021hypernerf, li2021neural, pumarola2021d}, and human movements~\citep{weng2022humannerf, peng2021neural}. 
    It is worth noting that the word, `dynamic', refers subjects' motions only. In this work, we extend the definition of the `dynamic' to varying illuminations as well as subjects' motions during taking an input video.

    Estimating and manipulating scene illuminations, such as intrinsic image decomposition~\citep{barrow1978recovering, horn1974determining}, relighting~\citep{debevec2000acquiring, xu2018deep} and shape-from-shading~\citep{zhang1999shape}, has been considered as one of classical research issues. There is a common assumption in these works that light sources are stable and predictable. Nevertheless, the main challenge of them is that the solution is not unique due to limited image resolution, noise and inaccurate camera geometry. To alleviate the challenges, proper prior information, such as depth geometry~\citep{chen2013simple, el2021ntire, maier2017intrinsic3d}, lighting directions~\citep{somanath2021hdr, wang2021learning} and segmentation masks~\citep{munkberg2022extracting}, is available, making their inferences more tractable.
        
    Despite the significant efforts of the previous works, there has been no attempt to explicitly handle varying illuminations in NeRFs. Unfortunately, NeRFs have no ability to learn neural fields under dynamic illuminations because the radiance of illumination can be considered as that of either static or dynamic objects, referred to \textit{illumination-radiance ambiguity}.
    
    As the first step toward addressing this issue, we need to find a proper prior to reduce the ambiguity of dynamic illuminations in neural fields. Our key observation is that the dynamic illuminations, intense and rapid lighting intensity and color changes during recording input videos, are created artificially in controlled situations, such as plays, shows and concerts. 
    As they often come with a rehearsal stage without the dynamic lighting effects before starting main stages, we use the video on the rehearsal stage as a prior.
    Although motions of the dynamic object in the rehearsal stage are not perfectly aligned with that of the main stage, it is valuable to represent scene geometry and to disentangle objects' own colors and the dynamic lighting effects in neural fields. 
 \begin{figure*}[t]
    \centering
     \includegraphics[width=\textwidth]{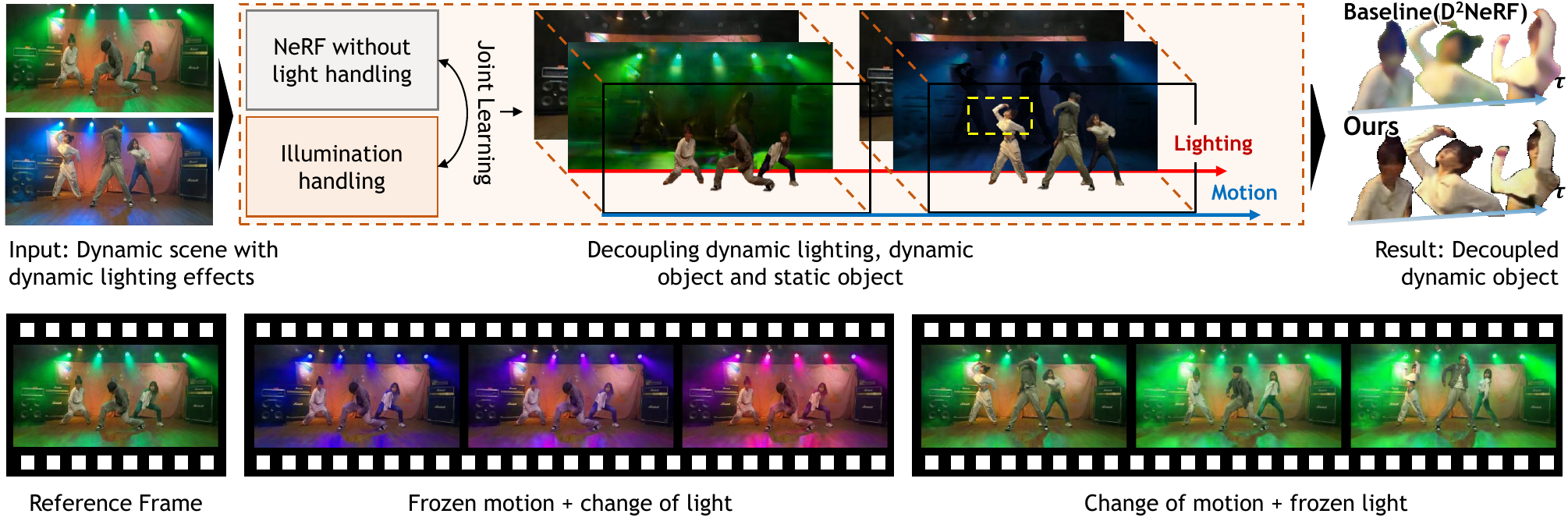}
     \caption{
     RehearsalNeRF jointly optimizes five neural fields for dynamic lighting and static/moving subjects in a training step.
     Each field cannot only be rendered independently, but can also be composited at once to represent a whole scene.
     Compared to the baseline~\citep{Wu2022d2nerf}, the details from RehearsalNeRF are distinguishable and the rendered colors are well decoupled from the scene lights.
     We provide a variety of applications for video editing, such as controlling lighting while stopping motion and vice versa.
     }
     \label{fig:teaser}
  \end{figure*}

     In this work, we present a \textbf{Rehearsal} prior-based \textbf{Ne}ural \textbf{R}adiance \textbf{F}ields (RehearsalNeRF) to synthesize and edit novel viewpoint images of a scene under dynamic lighting environments. 
    As shown in \Fref{fig:teaser}, our RehearsalNeRF learns to decouple dynamically changing illuminations and dynamic/static objects, simultaneously.
    To do this, we exploit the fact that scene geometry is consistent regardless of the lighting changes, except for regions of moving subjects, and the principle that a color is sum of emitted light energy from a surface.
    Our RehearsalNeRF first takes both rehearsal and main stage videos as inputs. With these inputs, we reconstruct three neural fields for static, dynamic objects and dynamically changing illuminations from lighting effects, respectively.

    To be specific, we develop a joint optimization for decoupling dynamic illuminations from moving/static objects in the scene.
    The main challenges of the joint optimization are two-fold: (1) optimizing dynamic illuminations is highly ill-posed due to the illumination-radiance ambiguity. Moreover, rapid illumination changes make learning radiance fields difficult; (2) learning a density field of dynamic objects is disturbed by shading changes from the illuminations as well.

    To solve the problems, we first introduce a learnable illumination vector to disentangle time-variant global and local illuminations of scenes. Different from existing appearance methods~\citep{martin2021nerf, tancik2022block}, it is an explicit representation with editable capability.
    We then find dense correspondences between a main stage and its rehearsal stage via optical flow estimation to account for color information of dynamic objects. Because optical flow is erroneous due to the illumination difference between them, we apply a regularizeration for the temporally consistent prediction.
    Finally, we design an additional regularizer using the rehearsal prior for robust supervision to decouple illuminations and static objects' own colors.
    
    We demonstrate the effectiveness of RehearsalNeRF by qualitatively and quantitatively evaluating it over state-of-the-art methods for NeRFs in the dynamic domain.
    In addition, our work is also applicable to areas of video manipulations including freezing motions under varying lights and vice versa.
\section{Related Works}\label{sec2}

\noindent\textbf{Dynamic neural scene representations}\quad
  There are three types of works categorizing dynamic neural scene representations: deformation of non-rigid parts, motions with human priors and decoupling dynamic part.
  
  To represent moving/deformable objects in a scene, D-NeRF~\citep{pumarola2021d} proposes neural implicit representations consisting of deformation fields and canonical fields to handle non-rigid motions. Nerfies~\citep{park2021nerfies} uses per-frame deformation latent codes, instead of time-stamps. HyperNeRF~\citep{park2021hypernerf} extends the Nerfies to account for topological changes using high-dimensional hyperspace. 
  For human motions, H-NeRF~\citep{xu2021h} and HumanNeRF~\citep{weng2022humannerf} use human body templates~\citep{alldieck2021imghum, loper2015smpl} to train the person in a canonical space. RigNeRF~\citep{athar2022rignerf} focuses on the deformation of head-pose and facial expressions using 3D morphable face model~\citep{blanz1999morphable}. Here, the template misalignment often has negative impacts on the reconstruction quality.
  To separate moving foregrounds from static backgrounds, several works have used priors such as optical flow~\citep{li2021neural, gao2021dynamic}, 3D depth~\citep{li2021neural, gao2021dynamic, xian2021space} and segmentation masks~\citep{zhang2021editable, tschernezki2021neuraldiff, li2022dynibar, jiang2022neuman}.
  Without any prior information, D$^2$NeRF~\citep{Wu2022d2nerf} decouples static and dynamic components, and handles shadow effects of moving objects in a self-supervised manner. NeRFPlayer~\citep{song2023nerfplayer} predicts sample-wise probabilities of being static, deforming and newness in a 4D spatio-temporal space.

  However, the previous works assume that a dynamic scene is under stable illumination conditions, and focus only on separating moving contents from static ones. In this case, the illumination-radiance ambiguity still remains, making them to confuse static objects as dynamic components.

\noindent\textbf{Neural representations for lighting and editing}\quad
  Thanks to the powerful 3D scene and light modeling capacities of neural implicit representations, there have been widely studied in intrinsic image decomposition~\citep{ye2023intrinsicnerf}, inverse rendering~\citep{munkberg2022extracting, boss2021nerd, boss2021neural, boss2022samurai, srinivasan2021nerv, physg2021, lyu2022neural, zhang2022iron} and shape-from-shading~\citep{ling2022shadowneus, yang2022s, zeng2023relighting}.
  Works in~\citep{munkberg2022extracting, boss2021nerd, boss2021neural, boss2022samurai, srinivasan2021nerv, physg2021, lyu2022neural} jointly optimize lighting, materials and scene geometry from multi-view images for neural representations using a differentiable renderer which is a learnable implementation of physically-based rendering (PBR) equation.
  Relighting~\citep{srinivasan2021nerv, lyu2022neural} with neural representations deals with global illuminations, 3D geometry and material information from a set of images with unconstrained known lighting locations.
  NeRV~\citep{srinivasan2021nerv} takes multiple images captured under known lighting conditions and produces a 3D representation of a scene, enabling to render novel viewpoint images with arbitrary lighting directions.
  ShadowNeuS~\citep{ling2022shadowneus} uses a shadow ray supervision to reconstruct neural signed distance fields from single-view images under multiple lighting conditions.
  IRON~\citep{zhang2022iron} improves a quality of geometry predictions and achieves the better relighting results by leveraging depth discontinuity on sampling. Recently, an inverse rendering done by optimizing 3D Gaussians~\citep{shi2023gir, liang2023gs} are proposed and shows faithful relighting results using PBR.

  We note that they only handle object-centric scenes with 360$^{\circ}$ observations. IntrinsicNeRF~\citep{ye2023intrinsicnerf}, which seems be similar with ours, optimizes editable neural representation of room-scale scenes. This decomposes a scene into reflectance and shading using a basic illumination model~\citep{phong1975illumination}, but does not handle time-variant lighting effects. 

\noindent\textbf{Appearance embedding for NeRF}\quad
  Following generative latent optimization~\citep{bojanowski2017optimizing}, many NeRF extensions use trainable latent codes for per-image appearance variations~\citep{martin2021nerf, turki2022mega, tancik2022block}, time-varying components~\citep{li2022neural,park2021nerfies, park2021hypernerf} and manipulation~\citep{schwarz2020graf, wang2022clip, niemeyer2021giraffe}.
  NeRF-in-the-wild~\citep{martin2021nerf} proposes a latent appearance modeling to address the scene inconsistency caused by different lighting conditions of unstructured photo-collections.
  Mega-NeRF~\citep{turki2022mega} and Block-NeRF~\citep{tancik2022block} handle the issue in large-scale scenes by embedding scene appearances like illumination changes according to camera poses into latent codes.
  DyNeRF~\citep{li2022neural} uses NeRF framework as a baseline and use temporal latent codes, allowing neural fields to be changed in time domain. CLIP-NeRF~\citep{wang2022clip} shows editable NeRF with text prompts to manipulate shapes and appearances of neural fields. GRAF~\citep{schwarz2020graf} and GIRAFFE~\citep{niemeyer2021giraffe} present 3D-aware controllable image synthesis using conditional neural fields. The objective of appearance embedding in the above methods is to facilitate matching with a canonical appearance for images under varying illuminations, yet cannot disentangle lighting effects from the scene. In contrast, our appearance embedding learns to explain dynamic lighting changes over time and shows the editable capability such as relighting and illumination color tune.

  \begin{figure*}[t]
    \centering
     \includegraphics[width=0.9\textwidth]{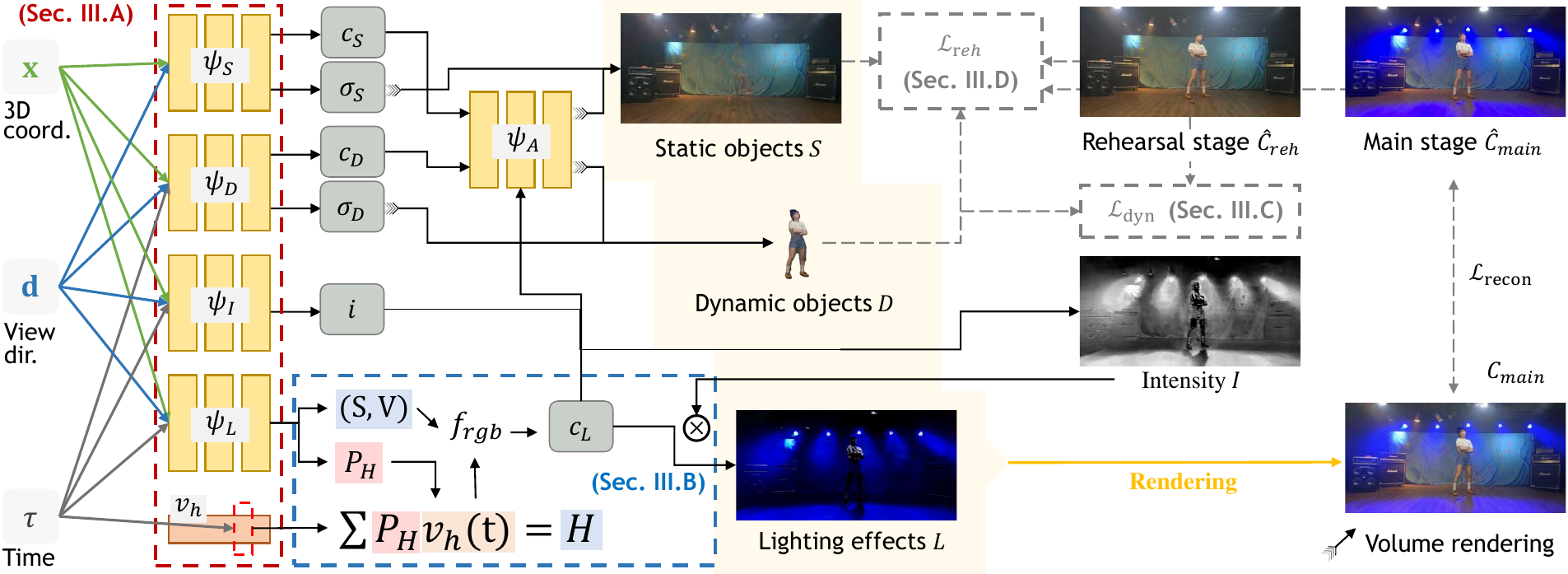}
     \caption{
     An overview of RehearsalNeRF in training phase. 
     Our RehearsalNeRF consists of five neural fields for rendering static, dynamic objects and illuminations.
     Each field takes location ($\mathbf{x}$), viewing direction ($\mathbf{d}$) and time-stamp ($\tau$) for the same sample point as input, except the time-stamp for the static fields. They are used to infer the radiance.
     Particularly, the illumination field predicts the probability $P_H$ of the illumination vector $v_h$ corresponding to $\tau$.
     To decouple static/dynamic objects and illumination components of the main stage video well, two additional regularizations, $L_{reh}$ and $L_{dyn}$, using the rehearsal prior, are designed.}
     \label{fig:overview}
  \end{figure*}

\section{Methodology}\label{sec3}

Given a camera ray $\mathbf{r}(t)=\mathbf{o}+t\mathbf{d}$ with near and far bound $t_n$ and $t_f$, respectively, where $\mathbf{o}$ is a ray origin and $\mathbf{d}\in R^3$ is a ray direction, NeRFs render view dependent colors $\mathbf{\tilde{C}}(\mathbf{r})$ as follows: 
\begin{multline}
     \mathbf{\tilde{C}} (\mathbf{r}) = \int_{t_n}^{t_f} T(t)\sigma(\mathbf{r}(t))\mathbf{c}(\mathbf{r}(t),\mathbf{d})~dt,~
     \text{s.t.}~~
     T(t) = \text{exp}\left(-\int_{t_n}^{t}\sigma(\mathbf{r}(s))ds\right),
    \label{eq: continuous volume rendering}
\end{multline}
where $\mathbf{c}(\mathbf{x},\mathbf{d})$ and $\sigma(\mathbf{x})$ denote a spatially-view-dependent radiance and a spatially-dependent density at spatial location $\mathbf{x}\in R^3$ with a depth $t$, respectively.
With the radiance fields of the scene built, we propose a novel neural scene representation pipeline which decouples components of dynamic lighting effects and objects, depicted in \Fref{fig:overview}. We refer to a scene with both dynamic illuminations and objects as \textit{main stage}, and without them as  \textit{rehearsal stage}. We first handle dynamic illumination change through separate neural fields (\ref{sec:scene_ref}) with a learnable illumination vector (\ref{sec:global_illum}). Then, we jointly optimize to resolve the illumination-radiance ambiguity on dynamic (\ref{sec:illumination}) and static objects (\ref{sec:prior}) with the proposed two regularizers.

  \begin{figure*}[t]
    \centering
     \includegraphics[width=0.8\textwidth]{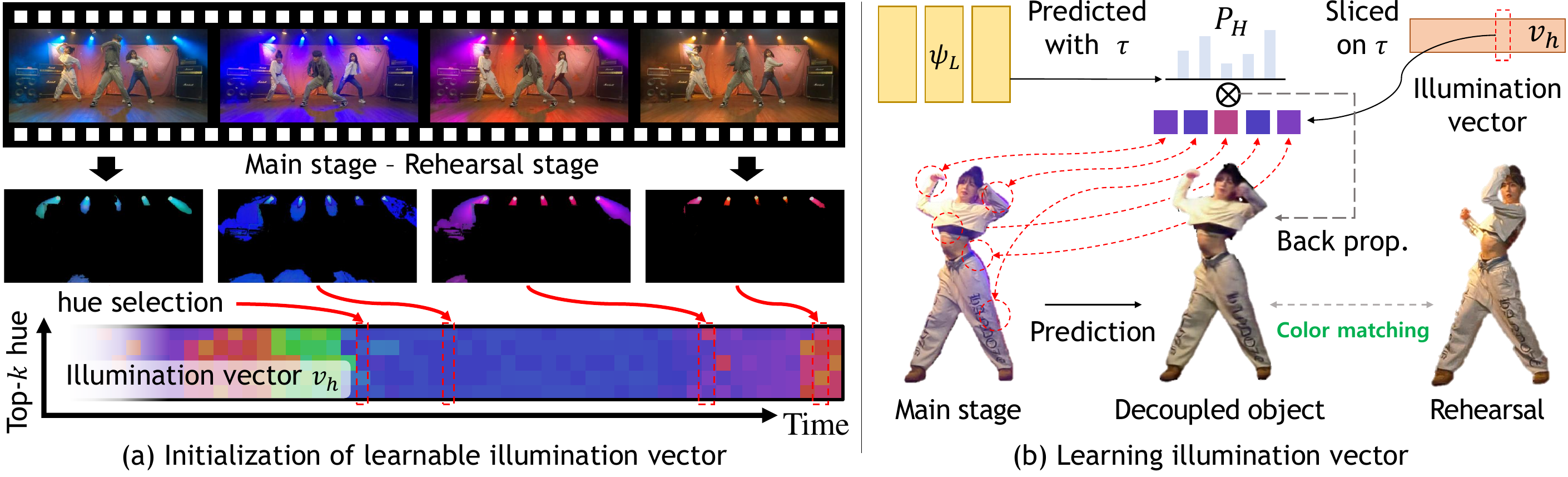}
     \caption{Visualizations of the procedure how to work the illumination vector.
     (a) The initialization of the learnable illumination vector is done by sampling the hue channel of the dynamic lighting from the difference between the rehearsal and the main stage video.
     (b) Back propagation from the rendered hue channel is achieved by computing a weighted sum of the probability $P_H$ and the slice of the $v_h$ for time $\tau$, which finally optimize the hue channel of the dynamic lighting effects.}
     \label{fig:illum_vector}
  \end{figure*}

\subsection{NeRF for Dynamic Illuminations}\label{sec:scene_ref}
For decoupling illuminations from static and dynamic objects each, we adopt D$^2$NeRF~\citep{Wu2022d2nerf} as a baseline, and compute a density $\sigma_S$ and a radiance $\mathbf{c}_S$ of static objects by using MLP layers $\psi_S$. We also calculate a density $\sigma_D$ and a radiance $\mathbf{c}_D$ of dynamic objects  from another MLP layers $\psi_D$. We thus obtain a radiance $\mathbf{c}_L$ of a dynamic illuminations by building multiple neural fields, whose overall process is denoted as function $\mathcal{F}_L$ below:
 \begin{multline}
   \psi_S : (\mathbf{x},\mathbf{d}) \rightarrow
   (\sigma_S,~\mathbf{c}_S),~
    \psi_D: (\mathbf{x},\mathbf{d},\tau) \rightarrow(\sigma_D,~\mathbf{c}_D),\\
        \psi_A: (\mathbf{c}_x,\mathbf{c}_L) \rightarrow
    (\boldsymbol{\alpha}_x),
    ~\mathcal{F}_L: (\mathbf{x},\mathbf{d},\tau) \rightarrow
    (\mathbf{c}_L),
    ~\text{s.t.}~~x \in [S,D],
\label{eq:D2nerf}
\end{multline}
where $\tau \in R$ is a temporal coordinate, and the MLP $\psi_A$ predicts a brightness of the illumination entangled with static and dynamic objects, $\boldsymbol{\alpha}_S \in R^3$ and $\boldsymbol{\alpha}_D \in R^3$, respectively.

 Here, we reformulate the volume rendering of NeRF~\citep{mildenhall2021nerf} as an explicit way to form editable radiances. We first introduce a learnable illumination vector ${v}_h$ for a self-supervision of the illuminations generated by light sources, and the function $\mathcal{F}_L$ to render time-varying illumination radiance $\mathbf{c}_L$, consisting of a neural field $\psi_L$. Details for the illumination vector and rendering the radiance of dynamic illumination $\mathbf{c}_L$ with $\mathcal{F}_L$ will be described in \ref{sec:global_illum}.

 The composite rendering of the baseline~\citep{Wu2022d2nerf} assumes that placing objects in the same position is physically impossible. However, the lighting effects exist on object surfaces in our problem setup. Considering this property, we modify the composite rendering equation for static and dynamic objects in the main stage by combining the radiances of illuminations as below\footnote{We omit writing $t$ for the simplicity from now}:
 
\begin{multline}
 \mathbf{C}_{main}^{S} = \int_{t_n}^{t_f} T'\sigma_S\beta(\mathbf{c}_S+\boldsymbol{\alpha}_S\odot\mathbf{c}_L) ~dt,
 ~
 \mathbf{C}_{main}^{D} = \int_{t_n}^{t_f} T'\sigma_D\beta(\mathbf{c}_D+\boldsymbol{\alpha}_D\odot\mathbf{c}_L) ~dt,
 \\ 
 \text{s.t.}~~
 T'(t) = \text{exp}\left(-\int_{t_n}^{t}(\sigma_S(\mathbf{r}(s))+\sigma_D(\mathbf{r}(s),\tau)) ~ds\right) 
 ~~\text{and}~~ \beta(x)=\frac{x}{x+1},
\label{eq: main rendering}
\end{multline}
where $\mathbf{C}_{main}^S$ and $\mathbf{C}_{main}^D$ are the rendered color of the static and dynamic objects in main stage, respectively. The $\beta$ is the tone-mapping operator~\citep{srinivasan2021nerv} and $\odot$ is the Hadamard operator.
In this equation, $\boldsymbol{\alpha}$ plays a role in a balancing term for the brightness of reflected rays in illuminations. The brightness $\boldsymbol{\alpha}$ depends on both radiance of objects and illuminations because an amount of the reflected light energy is closely related to the surface color. For example, a blue-colored surface reflects wavelength radiations for blue color when an incident light color is white. When a blue-colored light is projected onto a blue- and red-colored surface, the blue-colored surface becomes brighter while the red-colored surface absorbs the light. 

Based on \Eref{eq: main rendering}, the colors of the main stage $\mathbf{C}_{main}$ and lighting effects $\mathbf{C}_{main}^L$ can be rendered individually as follows:
\begin{multline}
    \mathbf{C}_{main}(\mathbf{r},\tau) = \mathbf{C}_{main}^S(\mathbf{r},\tau) + \mathbf{C}_{main}^D(\mathbf{r},\tau),
    \\
    \text{where}~~
    \mathbf{C}_{main}^{L}(\mathbf{r},\tau) = 
    \int_{t_n}^{t_f} T'\sigma_S\beta(\boldsymbol{\alpha}_S\odot\mathbf{c}_L) ~dt~
    +
    \int_{t_n}^{t_f} T'\sigma_D\beta(\boldsymbol{\alpha}_D\odot\mathbf{c}_L) ~dt.
\end{multline}
The color of static objects $\mathbf{\tilde{C}}_S$ and dynamic objects $\mathbf{\tilde{C}}_D$ without any dynamic illumination can be rendered based on 
\Eref{eq: continuous volume rendering} by using its corresponding density and radiance field from $\psi_S$ and $\psi_D$, respectively.

\subsection{Learning Illumination Vector}
\label{sec:global_illum}

Artificial lights in the main stage emit highly saturated colors, and the light energy varies according to a physical distance between the light source and its projected surface. In contrast, the hue channel of the light source is invariant to the distance. It means that the hue values are globally similar in the whole scene~\citep{zimmer2011optic}. Due to this property, we utilize HSV color space to represent the radiance field for lighting effects, instead of conventional RGB color space.

For hue values $H$, we formulate ${v}_h$ as a global, explicit and learnable component that contains candidate hue information of dynamic illuminations along with the time domain, as inspired by the concept of \cite{ye2023intrinsicnerf} and \cite{Meka2016live}. 
As shown in~\Fref{fig:illum_vector} (a), for effectively training ${v}_h$, we initialize the ${v}_h$ by taking centroids of $k$-means clustering~\citep{Meka2016live} of the hue values that have different colors between the rehearsal and the main stage scene.

\Fref{fig:illum_vector} (b) describes that the back propagation across the dynamic illumination rendering pipeline updates ${v}_h$. To render the dynamic illumination from the function in \Eref{eq:D2nerf}, we first build two neural fields, $\psi_{L}$ and $\psi_{I}$, to infer HSV color and intensity $i \in R$ of the dynamic illumination, respectively:
\begin{equation}{
    \psi_L: (\mathbf{x},\mathbf{d},\tau)\rightarrow
    (S,V,\mathbf{P}_{H},\sigma_L),~~
    \psi_I: (\mathbf{x},\mathbf{d},\tau) \rightarrow
    (i).
    }\label{eq3:two network}
\end{equation}
At last, we compute a saturation channel $S$, a value channel $V$ of the radiance and a probability vector $P_H$ from $\psi_{L}$, whose number of candidates is $k$. $P_H$ is used to calculate the hue channel $H$ corresponding to hue values of ${v}_h$ in a time $\tau$, and the color of dynamic illumination can be computed as below:
\begin{equation}{
    H =\sum_{j=1}^{k} \mathbf{P}_{H,j} {v}_{h,j}(\tau),\quad
    \mathbf{c}_L = f_{rgb}(H,S,V) \times i,
    }\label{eq3:HSV calcultaion}
\end{equation}
where ${v}_{h,j}(\tau)$ is a hue candidate at a specific time $\tau$ and hue index $j$. $\mathbf{P}_{H,j}$ is the probability corresponding to the ${v}_{h,j}(\tau)$, and $f_{rgb}$ is a function for converting HSV to RGB channels. The ${v}_{h}(\tau)$ is an explicit vector and capable of explaining spatially-invariant illumination colors. $\mathbf{P}_{H}$ contributes to continuous hue value expression across the image. In total, we can render the radiance of the dynamic illumination $\mathbf{c}_L$.

\subsection{Regularization for Illuminations on Dynamic Objects}\label{sec:illumination}

   As done in the baseline~\citep{Wu2022d2nerf}, decoupling dynamic and static components can be achieved with two separate neural radiance fields. However, it is infeasible to extract dynamic illuminations because the color changes of the scene do not depend on the illumination changes only, especially for dynamic objects by their motions.
   We address this problem by leveraging the illumination vector $v_h$ as a prior to identifying color of rays scattered by the light sources and the rehearsal prior.
   As shown in \Fref{fig:illum_vector} (b), $v_h$ facilitates to account for the illuminations on dynamic objects. If the hue values of pixels belonging to moving objects are similar to $v_h$, we can effectively render their colors, entangled with the ray colors, using knowledge of $v_h$.
   Then, the entangled colors can be effectively disentangled when the rehearsal prior provides colors of the dynamic objects as a coarse supervision.
   In consideration of this observation, we design a regularization $\mathcal{L}_{dynamic}$ which establishes a link of color information between the rehearsal prior and the dynamic object's neural field $\psi_D$ using optical flow~\citep{teed2020raft} $\mathcal{G}$ as follows:
\begin{multline}
    \mathcal{L}_{dyn} (\mathbf{r},\tau) 
    = 
    \mathcal{M}_{main}
    \lVert
    E_{reh \rightarrow main} \{ 
    \mathcal{G}_{reh \rightarrow main} (\mathcal{M}_{reh}\odot\mathbf{\hat{C}}_{reh}) - \mathbf{\tilde{C}}_D \}
    \rVert_2^2
    \\
    +\sum_{i=-2}^{2}
    \lVert
    E_{\tau+i \rightarrow \tau}^{dynamic} \{ \mathcal{G}_{\tau+i \rightarrow \tau}^{dynamic} (\mathbf{\tilde{C}}_D(\mathbf{r}, \tau+i))-\mathbf{\tilde{C}}_D(\mathbf{r}, \tau) \}
    \rVert_2^2,
\label{eq:dynamic reg}
\end{multline}
where $\mathcal{G}_{reh \rightarrow main}$ is the optical flow from rehearsal video to main stage video, and $\mathcal{G}_{\tau+i \rightarrow \tau}^{dynamic}$ is the optical flow between the rendered dynamic object at time $\tau + i$ and time $\tau$. $E$ refers to the confidence of the optical flow prediction~\citep{li2019learning}, $\hat{\mathbf{C}}_{reh}$ is a color of rehearsal video, and $\mathcal{M}_{main}$ and $\mathcal{M}_{reh}$ are binary masks of moving objects, predicted by an off-the-shelf interactive segmentation~\citep{cheng2021modular}, from main stage and rehearsal video, respectively.
To decouple the moving objects from the illuminations, this regularization cooperates with the color supervision in which the dynamic illumination is rendered:
\begin{equation}{
    \mathcal{L}_{recon} (\mathbf{r},\tau) =  \lVert \mathbf{\hat{C}}_{main} (\mathbf{r},\tau) - \mathbf{C}_{main} (\mathbf{r},\tau)
    \rVert_2^2,
}\label{Loss1:Recon loss}
\end{equation}
where $\mathbf{\hat{C}}_{main}(\mathbf{r},t)$ is the true color of the camera ray $\mathbf{r}$ at time $\tau$ in the main stage. The regularization in \Eref{eq:dynamic reg} also guides to learn the temporal coherency among the predicted colors of the dynamic objects. It helps the $\psi_D$ to overcome the illumination-radiance ambiguity because the dynamic objects' own colors are not changed significantly over time.

\subsection{Regularization for Illuminations on Static Objects}
\label{sec:prior}

Apart from the illumination vector $v_h$ for dynamic objects, illuminations on static objects should be considered. Here, we use the rehearsal prior one more time as in \Sref{sec:illumination}. Since the rehearsal prior provides information for scene geometry and subjects' original colors, we expect it to work in the static objects.

We train our radiance fields representing the static objects $\psi_S$ with the rehearsal prior based on a loss $\mathcal{L}_{reh}$ which is defined as below:
\begin{multline}
\mathcal{L}_{reh} (\mathbf{r},\tau) = 
(1-\mathcal{M}_{reh}) \lVert \mathbf{\hat{C}}_{reh} (\mathbf{r},\tau) - \mathbf{\tilde{C}}_{S} (\mathbf{r},\tau) \rVert_2^2
\\
+(1-\mathcal{M}_{main}) \lVert \mathbf{\hat{C}}_{main} (\mathbf{r},\tau) - \mathbf{C}_{main}^{S} (\mathbf{r},\tau) \rVert_2^2
\\
+(1-\mathcal{M}_{main}) \lVert \mathbf{\tilde{C}}_{S} (\mathbf{r},\tau) - (\mathbf{\tilde{C}}_{S} (\mathbf{r},\tau)+\mathbf{\tilde{C}}_{D} (\mathbf{r},\tau)) \rVert_2^2,
\label{Loss1:Rehearsal loss}
\end{multline}
where $\mathbf{\hat{C}}_{reh} (\mathbf{r},t)$ refers to the ground-truth color in the rehearsal stage.
The first and second terms in \ref{Loss1:Rehearsal loss} induce disentangling dynamic illuminations and static objects. The illumination-radiance ambiguity is resolved by leveraging the color of the rehearsal prior. The optimization of the illumination vector $v_h$ is conducted through rendering the main stage.
The third term mitigates the problem of neural fields not being learned in regions where $\mathcal{M}_{reh}$ and $\mathcal{M}_{main}$ do not overlap in the rendered image by providing a self-supervision for these regions. One important byproduct of this term is that, at this point, our method works in the absence of dynamic objects in the rehearsal stage. We can thus avoid a motion alignment issue between the main and rehearsal stage.

Lastly, the remaining issue is to disentangle static and dynamic objects. We address this issue by designing another regularization as below:
\begin{multline}
    \mathcal{L}_{p} (\mathbf{r},\tau) =
    \mathcal{M}_{main}\frac{\sigma_{S}(\mathbf{r},\tau)}{\sigma_{S}(\mathbf{r},\tau)+\sigma_{D}(\mathbf{r},\tau)} 
    + (1-\mathcal{M}_{main})\frac{\sigma_{D}(\mathbf{r},\tau)}{\sigma_{S}(\mathbf{r},\tau)+\sigma_{D}(\mathbf{r},\tau)}.
\end{multline}
This regularization penalizes static objects in density fields, and vise versa. Thanks to this, we can decouple them in the density fields.

\section{Experiments}
\subsection{Implementation Details}
 The neural fields ($\psi_S$, $\psi_D$ and $\psi_L$) of our RehearsalNeRF are implemented with a hybrid model of K-Planes~\citep{fridovich2023k} and D$^2$NeRF~\citep{Wu2022d2nerf}. Note that the neural fields can be replaced by other dynamic NeRF models.
 A training procedure of RehearsalNeRF requires a day for 120K iterations with 3,072 batch sizes on a single RTX 3090 GPU. We train our RehearsalNeRF with the following loss $\mathcal{L}$ :
\begin{multline}   
    \mathcal{L} (\mathbf{r},\tau) = \mathcal{L}_{recon} (\mathbf{r},\tau) + \lambda_{dyn}\mathcal{L}_{dyn} (\mathbf{r},\tau)  + \lambda_{reh}\mathcal{L}_{reh} (\mathbf{r},\tau) +\lambda_{p} \mathcal{L}_{p}(\mathbf{r},\tau).
\label{Loss3:overall loss}
\end{multline}

In our implementation, $\lambda_{dyn}$, $\lambda_{reh}$ and $\lambda_{p}$ are empirically set to 0.01, 0.5 and 0.01, respectively. In the experiments, we set the number of clustering of $k$-means algorithm to 5 because of the number of light sources in the main stages.

\subsection{Experiment Setup}
\noindent\textbf{Datasets.}\quad
 Since public datasets for dynamic NeRFs~\citep{li2022neural,broxton2020immersive} are limited to scenes captured under stable illuminations, we construct a video dataset for dance stages taken under dynamic colored lighting environments. Following the problem setting, our dataset contains pre-captured rehearsal videos under the stable lighting conditions before turning on the dynamic illuminations. Camera parameters are obtained from COLMAP~\citep{schoenberger2016sfm}. 
  We build a dataset consisting of synchronized multi-view videos whose spatial and temporal resolutions are 2704$\times$1520 and 120 FPS, respectively. For this, we use an array of 20 GoPro HERO cameras and take 10 video clips whose running time is 2.5 seconds.
  In the experiment, we downsample the spatial resolution by a factor of 2 and the frame rate to 30 FPS, similar to the previous work~\citep{li2022neural}. 

  As mentioned earlier, our work is the first attempt to handle dynamic illuminations in NeRF, and we cannot carry out an apple-to-apple comparison with publicly available neural rendering methods. Therefore, we have no choice but to conduct a limited evaluation of dynamic lighting effects. Unfortunately, it is infeasible to obtain the ground truth dynamic lighting effects which are highly entangled in the objects. As an alternative, we additionally collect a synthetic dataset simulated by Unity~\citep{juliani2018unity}. This dataset consists of 4 scenes whose number of frames is 30, and diverse lighting and motions exist. 

\begin{figure*}[t]
    \centering
     \includegraphics[width=\textwidth]{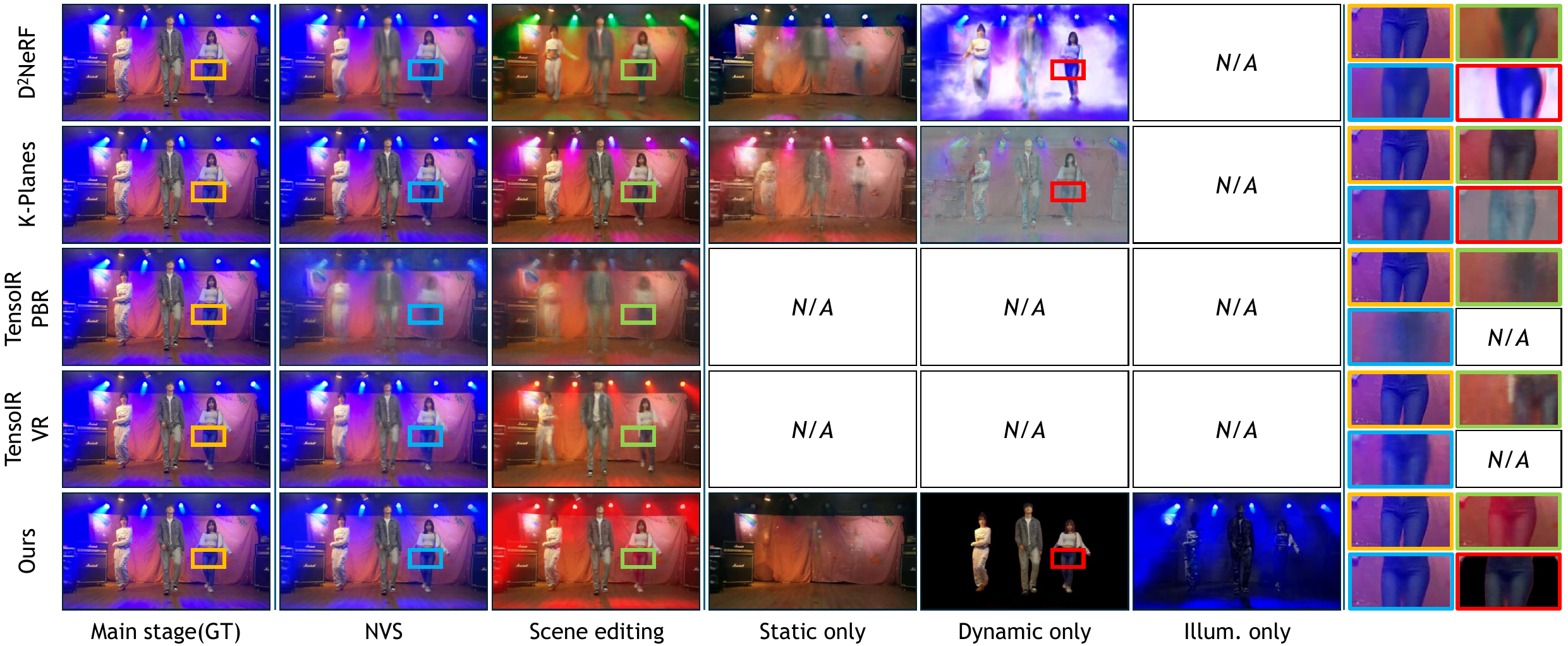}
     \caption{Qualitative comparisons on our real-world dataset. The goal of scene editing is to render the red-colored illumination.
  Only our method produces the high-quality edited scenes through the successful dynamic illumination decomposition.}
     \label{fig:realworld_editting}
\end{figure*}

\begin{table*}[t]
\centering
\caption{Quantitative comparison of novel view synthesis performance on our real-world dataset. We compare results from the original dataset and the temporally subsampled version (taking every $5^{th}$ frame) to evaluate robustness against rapid scene dynamics.}  
\label{tab:quantitative_realworld}

\resizebox{\linewidth}{!}{
    \begin{tabular}{l cc cc | cc cc}
        \toprule
        \multirow{3}{*}{\textbf{Method}} 
        & \multicolumn{4}{c|}{\textbf{Original Dataset}} 
        & \multicolumn{4}{c}{\textbf{Subsampled Dataset ($5\times$ Speed)}} \\
        
        \cmidrule(lr){2-5} \cmidrule(lr){6-9}
        
        & \multicolumn{2}{c}{\textbf{Full Image}} & \multicolumn{2}{c|}{\textbf{Dynamic Objects}} 
        & \multicolumn{2}{c}{\textbf{Full Image}} & \multicolumn{2}{c}{\textbf{Dynamic Objects}} \\
        
        & \textbf{PSNR}$\uparrow$ & \textbf{SSIM}$\uparrow$ & \textbf{PSNR}$\uparrow$ & \textbf{SSIM}$\uparrow$ 
        & \textbf{PSNR}$\uparrow$ & \textbf{SSIM}$\uparrow$ & \textbf{PSNR}$\uparrow$ & \textbf{SSIM}$\uparrow$ \\
        \midrule
        
        D$^2$NeRF~\citep{Wu2022d2nerf} 
        & \textbf{29.776} & 0.857 & 22.962 & 0.978 
        & 28.274 & \underline{0.853} & 21.899 & 0.976 \\
        
        K-Planes~\citep{fridovich2023k}
        & 28.113 & \textbf{0.867} & \underline{25.122} & \underline{0.984} 
        & 27.849 & \textbf{0.854} & \underline{25.784} & \underline{0.986} \\
        
        TensoIR PBR~\citep{jin2023tensoir} 
        & 18.681 & 0.707 & 16.934 & 0.972 
        & 16.955 & 0.708 & 15.347 & 0.971 \\
        
        TensoIR VR~\citep{jin2023tensoir} 
        & \underline{28.663} & \underline{0.865} & 21.918 & 0.978 
        & \underline{28.389} & 0.848 & 23.238 & 0.981 \\
        
        4DGaussian~\citep{wu20244d} 
        & 28.199 & 0.851 & 24.523 & \underline{0.984} 
        & 28.189 & 0.842 & 24.245 & 0.983 \\
        
        \midrule
        
        \textbf{Ours}
        & 28.250 & 0.851 & \textbf{26.318} & \textbf{0.986} 
        & \textbf{28.666} & 0.845 & \textbf{27.090} & \textbf{0.988} \\
        \bottomrule
    \end{tabular}
}
\end{table*}

\noindent\textbf{Comparison methods.}\quad
For strictly fair comparison, we modify comparison methods for handling dynamic illuminations as below:

\noindent(1) D$^2$NeRF~\citep{Wu2022d2nerf} disentangles dynamic and static objects. As in \cite{park2021hypernerf}, using it as a baseline, we apply an appearance latent vector to embed the illumination environment. The latent vector allow us to interpolate and manipulate the lighting changes.

\noindent(2) K-Planes~\citep{fridovich2023k} consists of six planes to represent dynamic motions. By adding time-variant appearance embedding to their 4D dynamic volume representation, we can manipulate the time-variant illumination.

\noindent(3) TensoIR~\citep{jin2023tensoir} also introduces an appearance tensor, so we can learn and infer the time-varying illumination and editing. We adopt the same training scheme with \cite{chen2022relighting4d}. TensoIR is optimized by using physically based rendering (PBR) with the help of volume rendering (VR). Both results are reported in this evaluation.

\noindent(4) 4DGaussian~\citep{wu20244d} leverages a 4D voxel field to decode temporal features for 3DGS parameters, enabling dynamic scene representation. However, it does not separate dynamic illumination from dynamic objects. To verify the impact of this limitation, we measure its novel view synthesis performance, focusing on how well it reconstructs dynamic objects under dynamic illuminations.

  \begin{figure*}[t]
    \centering
     \includegraphics[width=\textwidth]{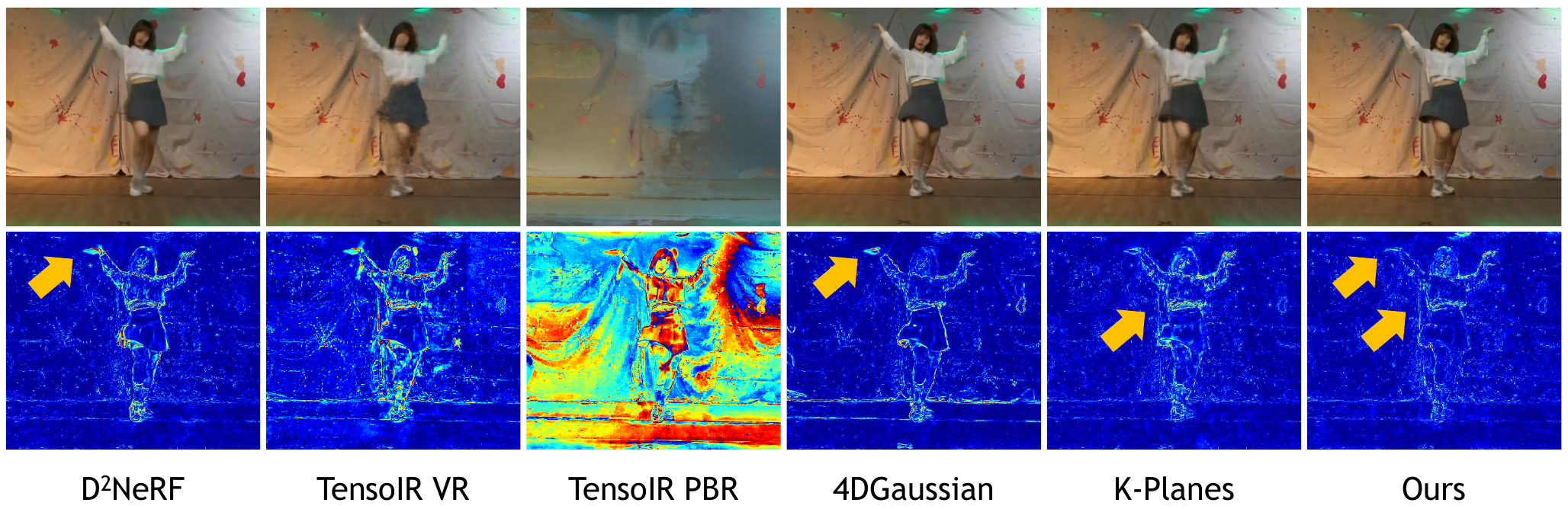}
     \caption{
     Qualitative comparison of novel view synthesis performance for dynamic objects on our real-world dataset. The second row presents error map between rendered image and GT. Our model consistently achieves superior quality in dynamic regions.
     }
     \label{fig:NVS_on_dynamic}
  \end{figure*}

\subsection{Results for Novel View Synthesis}
\label{sec:evaluation}

We evaluate the novel view synthesis (NVS) performance of RehearsalNeRF using PSNR and SSIM~\citep{ssim} as quantitative metrics. Evaluations are conducted on both our real-world dataset and a synthetic dataset that simulates dance stages with complex illumination.

We first evaluate ReheasrsalNeRF on our real-world dataset in in \Tref{tab:quantitative_realworld}. Regarding the results, D$^2$NeRF and K-Planes exhibit the higher PSNR and SSIM values than ours. That is because of their overfitting issue. They directly learn highly entangled color representations without considering the illumination–radiance ambiguity over time, and the naive appearance embeddings used in these methods are insufficient to disentangle the radiance fields of dynamic objects from illumination. TensoIR with PBR fails because the illumination change disturbs learning scene geometry. Meanwhile, TensoIR with VR renders plausible results because it also overfits to the illumination changes in the radiance field. 
Consequently, this overfitting degrades their ability to accurately reconstruct dynamic objects (see \Fref{fig:NVS_on_dynamic}) and further limits their scene-editing capabilities (as demonstrated in \Fref{fig:realworld_editting}).

\begin{figure*}[t]
    \centering
    \includegraphics[width=\textwidth]{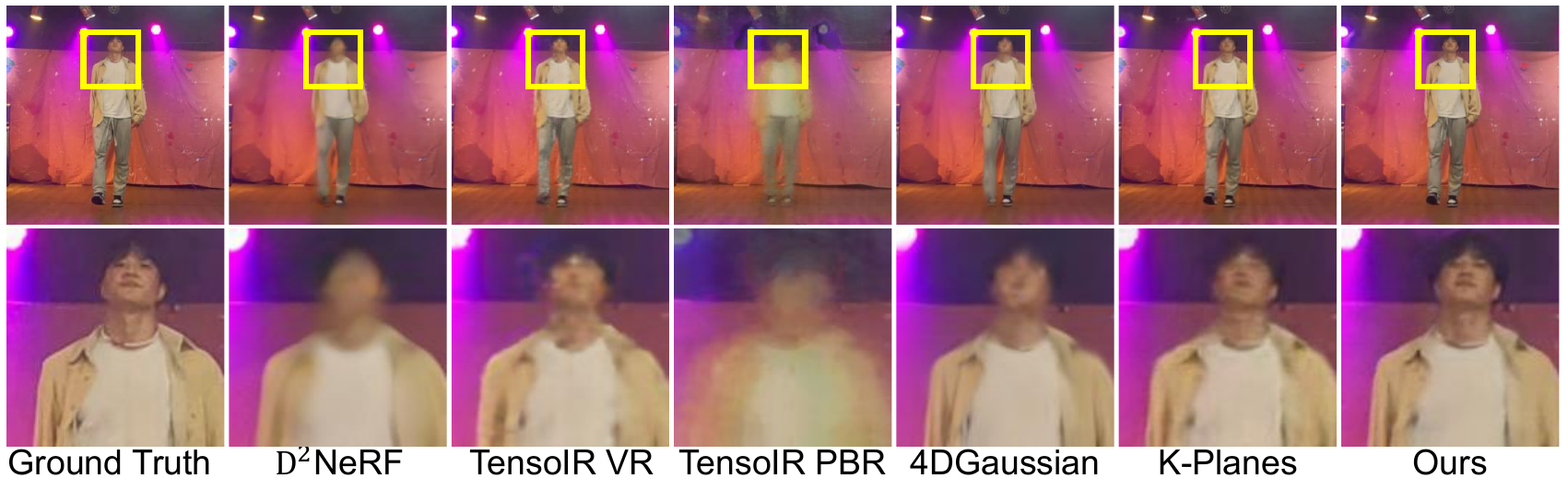}
    \caption{
    Qualitative comparison of NVS performance for dynamic objects on the temporally subsampled real-world dataset. Due to rapid motion, baselines struggle to reconstruct the fine details of dynamic objects.
    }
    \label{fig:NVS_5frame_dynamic}
\end{figure*}

To verify that the baselines' high performance stems from overfitting rather than true robustness, we conducted further evaluations on more dynamic scenarios. We constructed a temporally subsampled dataset by taking every $5^{th}$ frame from the original videos, effectively simulating rapid motion ($5\times$ speed). While existing approaches lack the capability to handle such rapid dynamics, our pipeline is specifically designed to disentangle dynamic lighting from dynamic objects. As reported in \Tref{tab:quantitative_realworld}, although K-Planes maintains a marginal lead in SSIM for full images, our method achieves the highest PSNR across all compared models. Crucially, as scene dynamics increase, competing methods suffer significant degradation due to severe color inconsistencies. This trend is particularly evident in the dynamic object evaluation, where our method demonstrates superior fidelity. This observation is qualitatively corroborated by \Fref{fig:NVS_5frame_dynamic}, which illustrates the loss of fine details in dynamic objects produced by the baselines.

\begin{figure*}[t]
    \centering
     \includegraphics[width=\textwidth]{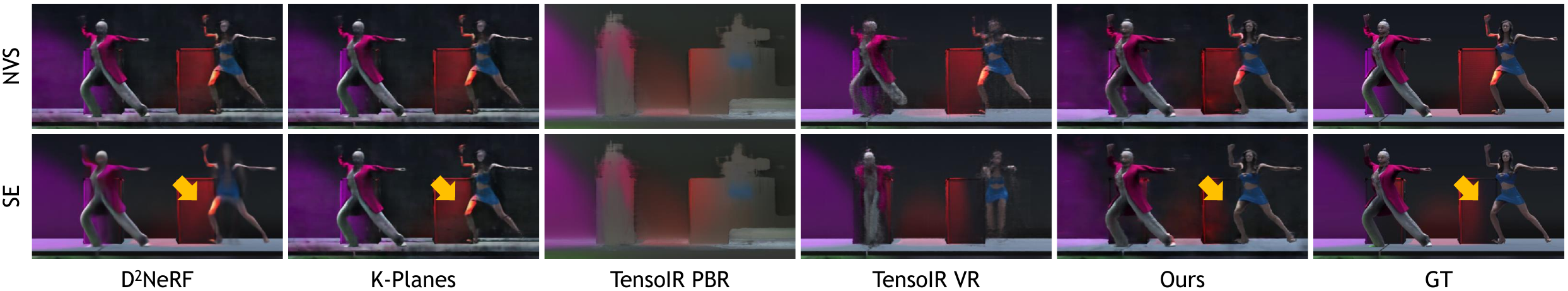}
     \caption{Qualitative comparison for scene editing on the synthetic dataset. The comparison methods fail to handle the harsh illuminations, mistakenly optimizing dynamic lights as dynamic objects.}
     \label{fig:comparison}
  \end{figure*}

\begin{figure*}[t]
    \centering
     \includegraphics[width=\textwidth]{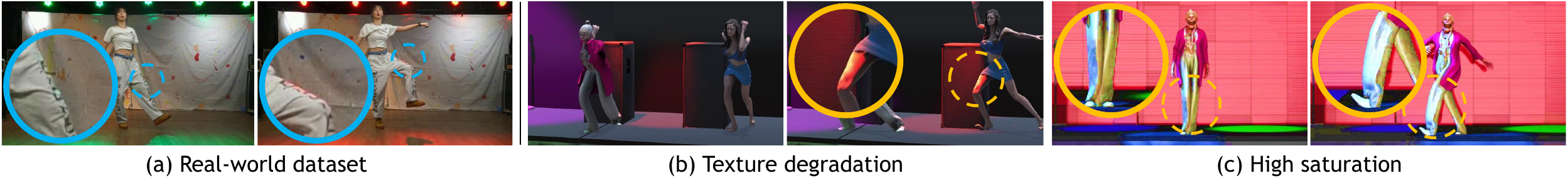}
     \caption{Examples from our real-world (a) and synthetic datasets (b,c). The synthetic dataset allows us to test ours and the comparison models in severe illumination conditions, including (b) texture degradation and (c) highly saturated regions.}
     \label{fig:dataset_eg}
  \end{figure*}

Our robustness is further demonstrated on our synthetic dataset in \Tref{tab:quantitative_synthetic} and~\Fref{fig:comparison}, which features severe lighting conditions. Although the videos suffer from the degradation of the objects' textures (\Fref{fig:dataset_eg} (b)) and saturated colors (\Fref{fig:dataset_eg} (c)), our disentanglement technique operates to characterize the extent of temporal changes of pixels related to the lighting effects. Please refer to our Appendix C.2 for more analysis.

\begin{table*}[t]
\centering
\small
\caption{Quantitative evaluation for novel view synthesis and scene editing on our synthetic dataset with state-of-the-art methods. To evaluate scene editing performance, we construct a synthetic video dataset with time-shifted lighting effects relative to the original video used as the learning objective.}
\label{tab:quantitative_synthetic}

\resizebox{0.8\linewidth}{!}{
    \begin{tabular}{l cc cc}
        \toprule
        \multirow{2}{*}{\textbf{Method}} & \multicolumn{2}{c}{\textbf{Novel View Synthesis}} & \multicolumn{2}{c}{\textbf{Scene Editing}} \\
        
        \cmidrule(lr){2-3} \cmidrule(lr){4-5}
        
         & \textbf{PSNR} $\uparrow$ & \textbf{SSIM} $\uparrow$ & \textbf{PSNR} $\uparrow$ & \textbf{SSIM} $\uparrow$ \\
        \midrule
        
        D$^2$NeRF~\citep{Wu2022d2nerf} 
        & 27.234 & 0.872 & 25.201 & 0.864 \\
        
        K-Planes~\citep{fridovich2023k} 
        & 27.495 & 0.864 & 25.783 &0.853 \\
        
        TensoIR PBR~\citep{jin2023tensoir} 
        & 15.056 & 0.668 & 15.153 & 0.671 \\
        
        TensoIR VR~\citep{jin2023tensoir} 
        & \underline{28.769} & \textbf{0.885} & \underline{27.762} & \textbf{0.883} \\
                
        \midrule
        
        \textbf{Ours} 
        & \textbf{30.117} & \underline{0.883} & \textbf{29.186} & \underline{0.878} \\
        \bottomrule
    \end{tabular}
}
\end{table*}

\subsection{Results for Scene Editing}
\label{sec:quality on real world}
\noindent\textbf{Quantitative comparison.}\quad
Our ultimate goal of this work is to show the capability of the explicit representation for dynamic illuminations. To demonstrate it, we propose an exclusive evaluation protocol that generates lighting effects of videos in time from its 10-frame shifted version. The scene editing evaluation is conducted on our synthetic dataset.

For strictly fair comparison, we replace the appearance time code for D$^2$NeRF and K-Planes, which is embedded in a training phase. For TensoIR, we make the shifted version of the appearance tensor by 10 frames in a lighting dimension. 

In \Tref{tab:quantitative_synthetic}, our method outperforms them. If the illumination is incorrectly decoupled, the predicted colors of objects differ from their original colors, which have a negative impact on the PSNR values of the comparison methods.

\noindent\textbf{Qualitative comparison.}\quad
 \Fref{fig:realworld_editting} shows the qualitative comparisons on our real-world dataset for scene editing and its corresponding scene decomposition. 
 For scene editing, we extract an appearance embedding from a red-colored illumination in the main stage video frame, and use it as the appearance vector.

As illustrated in \Fref{fig:realworld_editting}, the comparison methods, adopting implicit representation for dynamic components, face the challenges in lighting color synthesis. Furthermore, the scene decomposition into simple dynamic and static objects suffers from rendering them under the entangled illuminations. In particular, they fail to represent spread lighting effects on dynamic objects.

In contrast, our RehearsalNeRF demonstrates the remarkable capabilities in both scene editing and decomposition. In the synthetic dataset, as shown in \Fref{fig:comparison}, ours reveals the significant performance gap from the comparison methods.

\begin{table*}[t]
    \centering
    \caption{Quantitative comparison of dynamic lighting disentanglement on our synthetic dataset. We report L2 loss, L1 loss, and PSNR.}
    \label{tab:disentanglement_metrics} 
    \resizebox{0.6\linewidth}{!}{
            \begin{tabular}{lccc}
            \toprule
            \textbf{Method} & \textbf{L2} ($\times 1000$) $\downarrow$ & \textbf{L1} ($\times 1000$) $\downarrow$ & \textbf{PSNR} $\uparrow$ \\
            \midrule
            K-Planes        & 14.897 & 81.276 & 19.523 \\
            D$^2$NeRF       & 3.147  & 17.811 & 25.245 \\
            \midrule
            \textbf{Ours}   & \textbf{0.398} & \textbf{7.630} & \textbf{34.972} \\
            \bottomrule
        \end{tabular}
    }
\end{table*}
\begin{figure*}[t]
    \centering
     \includegraphics[width=\textwidth]{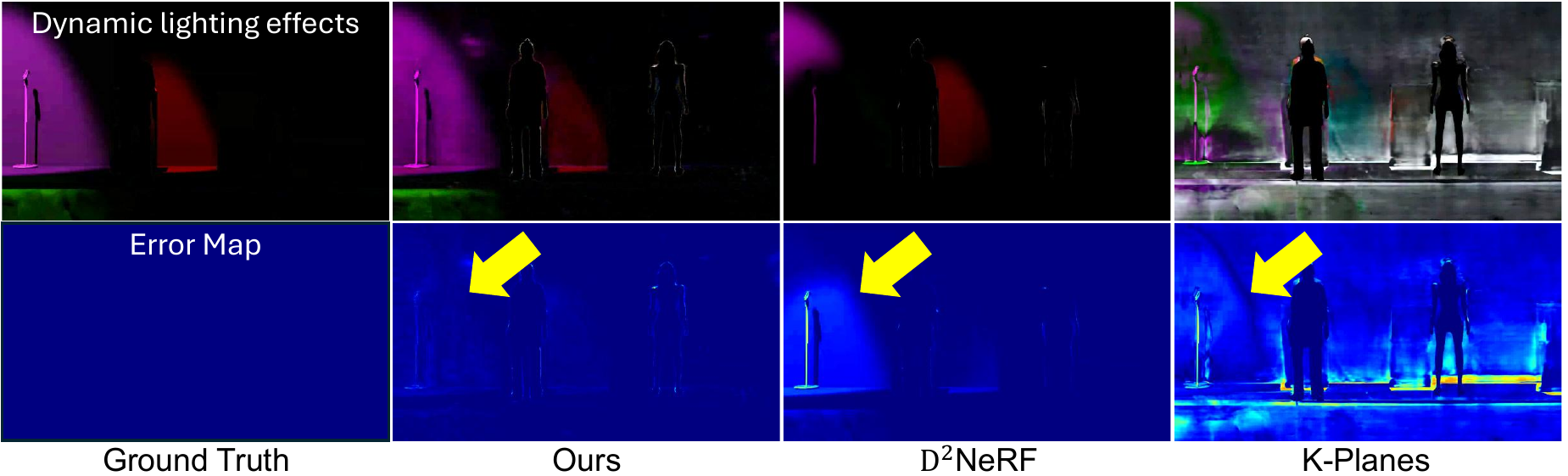}
     \caption{
     Qualitative comparison of dynamic lighting disentanglement.
     }
     \label{fig:decoupling_light}
  \end{figure*}

\noindent\textbf{Dynamic light disentanglement}\quad
To evaluate the quality of dynamic lighting disentanglement, as shown in Ground Truth of \Fref{fig:decoupling_light}, we obtain ground truth dynamic lighting effects maps by calculating the difference between images from the main and rehearsal stages, while masking out dynamic objects. For the baselines, we extract dynamic lighting effects by subtracting the rendered static parts from the full rendered images. We then quantify the performance by measuring the errors between the ground truth maps and the estimated lighting effects from the baselines. The quantitative and qualitative results are demonstrated in \Tref{tab:disentanglement_metrics} and \Fref{fig:decoupling_light}, respectively.

\begin{figure*}[t]
    \centering
     \includegraphics[width=0.95\textwidth]{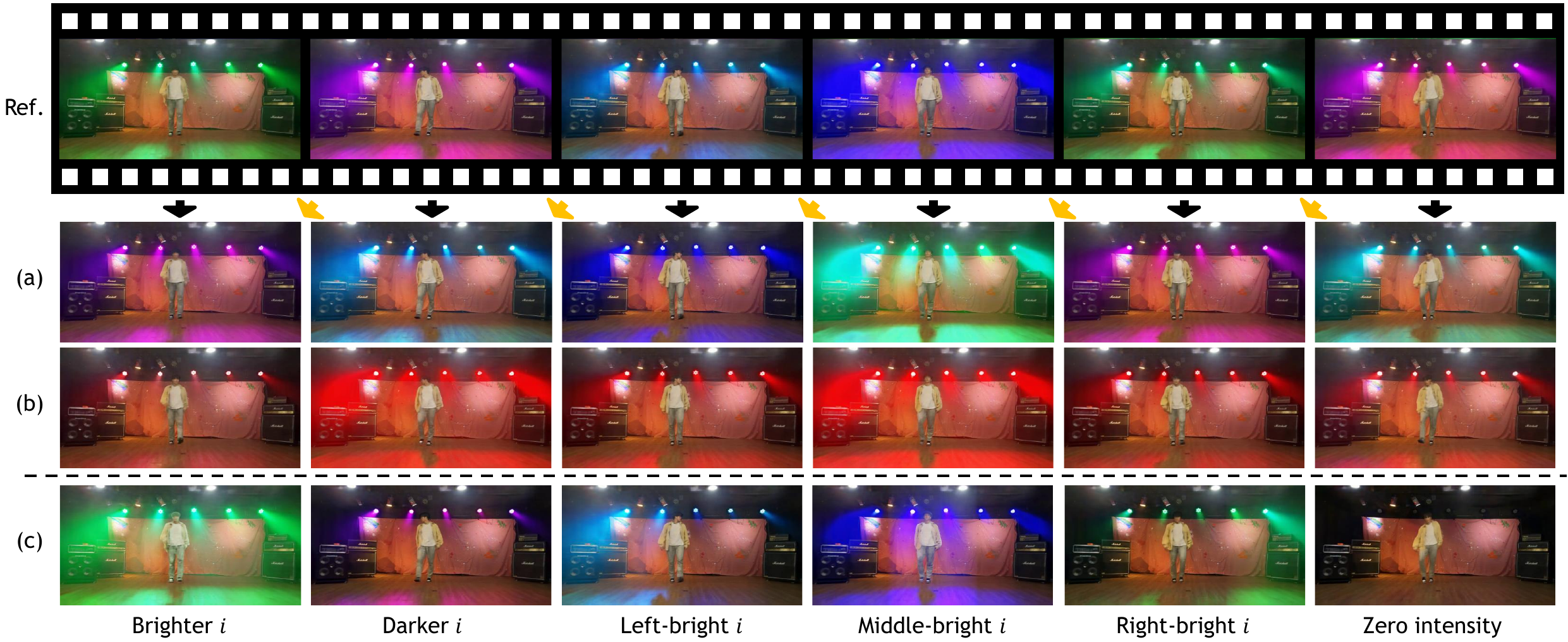}
     \caption{Qualitative results for scene editing. (a) We shift the lighting effects to the former frame (see yellow allows). Note that, we do not modify the intensity of illuminations at this point. (b) We tune $v_h$ with the red-colored illumination. (c) An impact on the modified intensity value $i$. We can infer the brighter and darker illumination by applying the lower and higher value of $i$, respectively. Plus, we can implement the spatially-variant adjustment.}
     \label{fig:edit}
  \end{figure*}

\noindent\textbf{Application.}\quad 
RehearsalNeRF allows to control scene components like subjects' motions and illuminations using our novel explicit and continuous representation. In \Fref{fig:edit} (a), we can change the time zone of the lighting by tuning $\tau$ on the illumination vector $v_h$. We can also manipulate the illumination colors by manually defining colors of $v_h$, whose example is in \Fref{fig:edit} (b). As shown in \Fref{fig:edit} (c), we can modulate the light brightness with the intensity value $i$ of $\psi_I$, and artificially make the spatially-varying intensity by assigning new factors to the $i$.

In the Appendix A, we also show an interesting application to confocal fluorescence microscopy, which makes novel viewpoint images on both cells and tissues captured under multiple light sources with different wavelengths.

\begin{figure*}[t]
    \centering
     \includegraphics[width=0.95\textwidth]{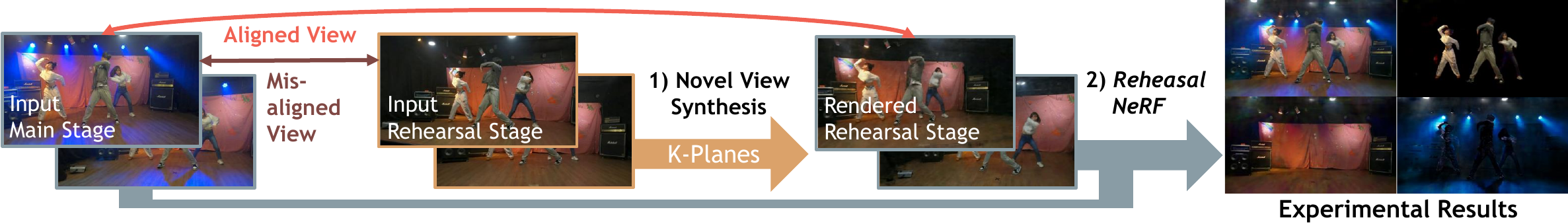}
     \caption{Handling a large misalignment of cameras between the main stage and the rehearsal stage.}
     \label{fig:Cam_misalign}
  \end{figure*}
 
\begin{figure*}[t]
\centering\includegraphics[width=0.96\textwidth]{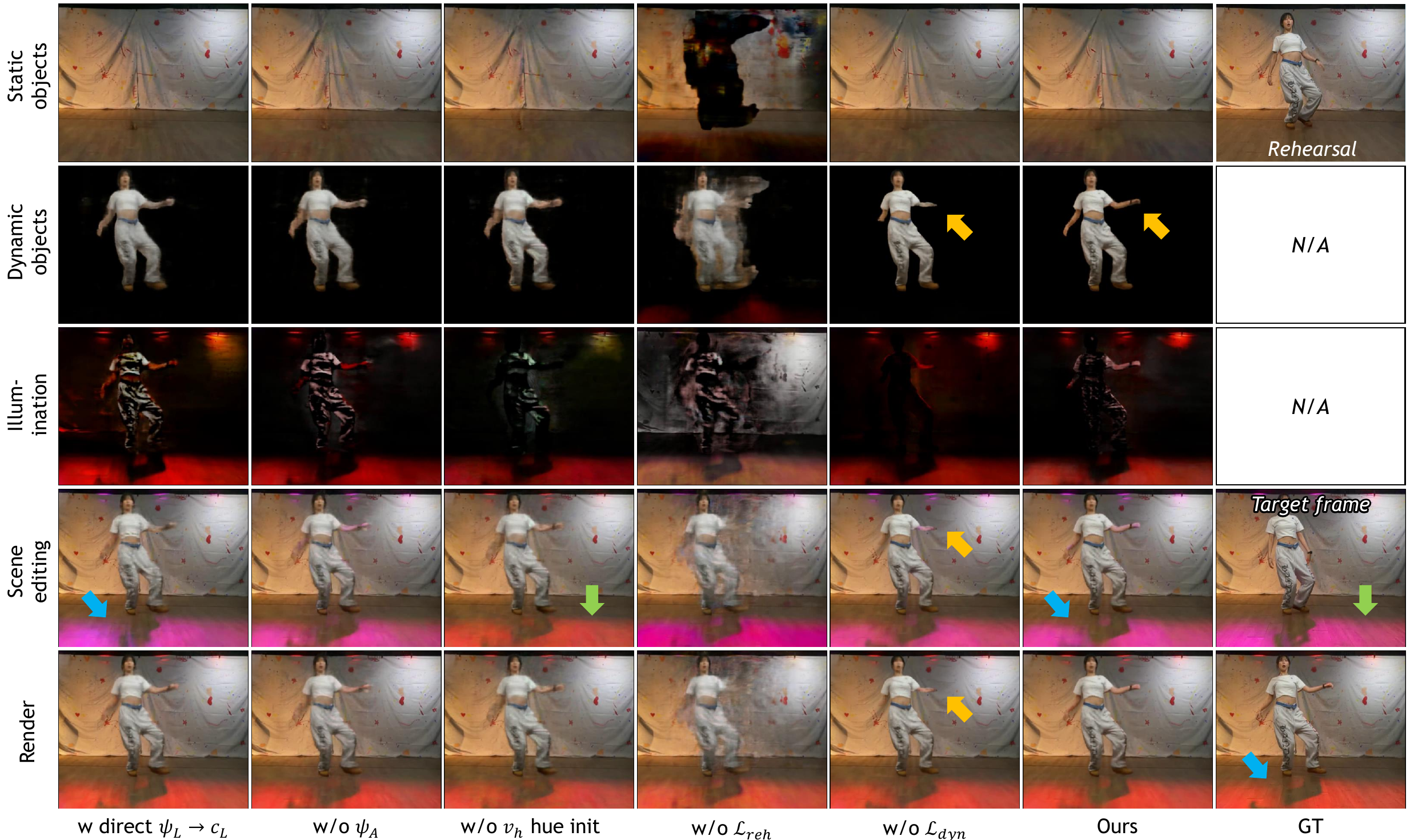}
 \caption{Visualizations of the results on ablation study. We can see that each component of our RehearsalNeRF contributes to the visually pleasing looks. The dynamic object disappearance (yellow arrow), incorrect scene editing (green arrow) and inaccurate shadow (blue arrow) are observed if we do not have either one of the proposed components.}
\label{fig:ablation}
\end{figure*}

\subsection{Ablation Study}
\label{sec:ablation}

In~\Tref{tab:ablation} and~\Fref{fig:ablation}, we demonstrate the effectiveness of the regularization $\mathcal{L}_{dyn}$, $\mathcal{L}_{reh}$, the hue initialization for illumination vector $v_h$, $\psi_A$, and our novel rendering procedure of light colors $c_L$ in \Sref{sec:global_illum}. Each component contributes to the performance of our RehearsalNeRF, whose analyses are described below:

\noindent(1) Without $\mathcal{L}_{dyn}$, SSIM values slightly increase due to optical flow confidence predictions, where reliable pixels are sparsely distributed. It offers a limited supervision of rehearsal prior. Although the radiance field attempts to recover a continuous signal, the discrete nature of rehearsal prior observations can degrade the SSIM values. Conversely, PSNR values may improve as it focuses on minimizing the mean-squared error in ours.

\noindent(2) Learning $\psi_A$ enables the accurate color representation when illuminations have a impact on surface appearances. Conversely, omitting $\psi_A$ results in the lower PSNR but higher SSIM because static objects are dominant in a scene. Static objects show color variations solely caused by dynamic illuminations, whereas dynamic objects are influenced by both their motions and dynamic illuminations. Therefore, for static objects, optimizing $\psi_A$ may be unnecessary, as the continuous representation tends to yield smooth results, which might reduce the SSIM values.

\noindent(3) We can compute the illumination color $c_L$ directly from the neural field $\psi_L$, bypassing the method in \Sref{sec:global_illum}. It unexpectedly achieves the higher PSNR values for real-world novel view synthesis. However, this may indicate a potential overfitting issue where $\psi_L$ learns the entangled lighting colors on object surfaces. They also lack scene editing capabilities demonstrated in \Sref{sec:quality on real world} because they are able to only manipulate the appearance embedding.

\begin{table*}[t]
\centering
\small
\caption{Ablation study for NVS on both real-world and synthetic dataset, and SE.
}\label{tab:ablation}
\resizebox{0.8\textwidth}{!}{
\begin{tabular}{@{}l@{}||@{}c@{}c@{}|@{}c@{}c@{}|@{}c@{}c@{}}
\Xhline{2\arrayrulewidth}
& \multicolumn{2}{c|@{}}{~~NVS(Real-world)~~} & \multicolumn{2}{c|}{~~NVS(Synthetic)~~} & \multicolumn{2}{c}{~~SE(Synthetic)~~} \\ 
& ~~~~PSNR~~~~ & SSIM~~~~ & ~~~~PSNR~~~~ & SSIM~~~~ & ~~~~PSNR~~~~ & SSIM~~~~ \\ \hline
 
\multirow{1}{*}{Ours}
& \underline{28.250} & \underline{0.851}~~~~ & \bf{30.117} & \underline{0.883}~~~~ & \bf{29.186} & \underline{0.878}~~~~ \\

\multirow{1}{*}{w/o $\mathcal{L}_{dyn}$~~~~}
& 27.787 & \bf{0.859}~~~~ & 29.769 & 0.879~~~~ & 25.948&  0.859~~~~ \\ 

\multirow{1}{*}{w/o $\mathcal{L}_{reh}$~~~~}
& 27.845 & 0.827~~~~ & 25.896 & 0.852~~~~ & 25.190 & 0.846~~~~ \\ 

\multirow{1}{*}{w/o $v_h$ hue init.}
& 27.378 & 0.832~~~~ & 29.439 & 0.878~~~~ & 28.574 & 0.872~~~~ \\

\multirow{1}{*}{w/o $\psi_A$}
& 28.229 & 0.845~~~~ & \underline{30.040} & \bf{0.885}~~~~ &\underline{29.104} & \bf{0.880}~~~~ \\ 
 
\multirow{1}{*}{w direct $\psi_L \rightarrow c_L$~}
& \bf{28.345} & 0.816~~~~ & 28.908 & 0.871~~~~ & 27.820 & 0.864~~~~ \\ 

\Xhline{2\arrayrulewidth}
\end{tabular}}
\end{table*}

\subsection{Analysis}
\label{sec:analysis}

\begin{table*}[t]
\centering
\small
\caption{Quantitative result for experiments with time mis-sync rehearsal videos.
}
{
\resizebox{0.8\textwidth}{!}{

\begin{tabular}{@{}c@{}c@{}|@{}c@{}c@{}|@{}c@{}c@{}|@{}c@{}c@{}}
\Xhline{2\arrayrulewidth}
\multicolumn{2}{c|@{}}{~~Ours~~} & \multicolumn{2}{c|}{+30 frame shift~~} & \multicolumn{2}{c|}{+60 frame shift~~} & \multicolumn{2}{c}{Only use 1$^{st}$ frame~~} \\ 
~~~~PSNR~~~~ & SSIM~~~~ & ~~~~PSNR~~~~ & SSIM~~~~ & ~~~~PSNR~~~~ & SSIM~~~~ & ~~~~PSNR~~~~ & SSIM~~~~ \\ \hline
 
\underline{28.046} & \bf{0.883}~~~~ &
28.039 & \underline{0.882}~~~~ &
\bf{28.069} & \bf{0.883}~~~~ &
28.005 & 0.881~~~~ \\

\Xhline{2\arrayrulewidth}
\end{tabular}}
}
\label{tab:time_missync}
\end{table*}

\begin{figure*}[t]
    \centering
     \includegraphics[width=0.95\textwidth]{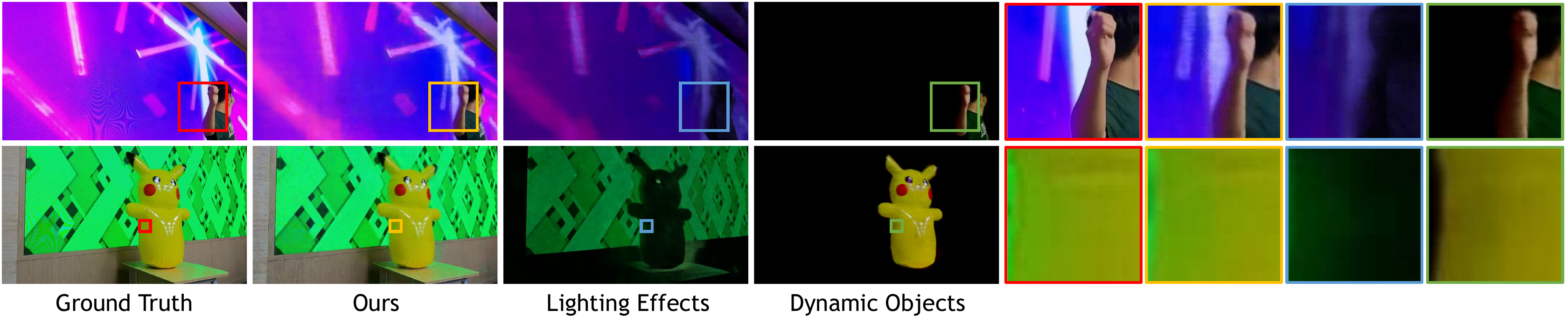}
     \caption{Experimental results when the subjects are captured with the bulky LED screen. Adjusting the hyper-parameter $k$ makes it possible to represent complex lighting effects, such as those from an LED screen.}
     \label{fig:LED}
  \end{figure*}

\begin{figure*}[t]
    \centering
     \includegraphics[width=0.95\textwidth]{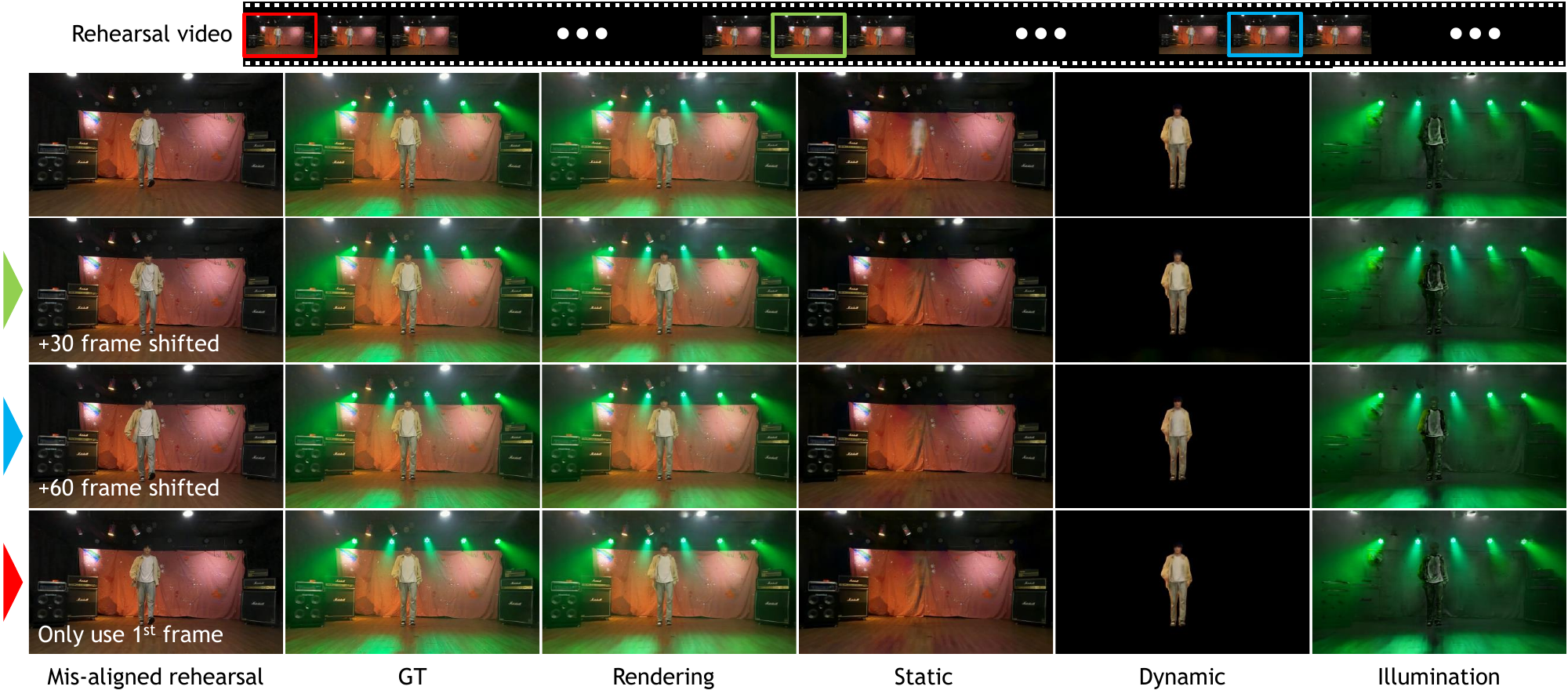}
     \caption{Experimental results with non-synchronized rehearsal video input. The time shift between the rehearsal and main stage videos results in a misalignment of dynamic motions. Using only the first frame of the rehearsal video as the rehearsal prior exacerbates this misalignment. Our RehearsalNeRF remains unaffected by such misalignments.}
     \label{fig:mis_sync}
  \end{figure*}

\begin{figure*}[t]
    \centering
     \includegraphics[width=\textwidth]{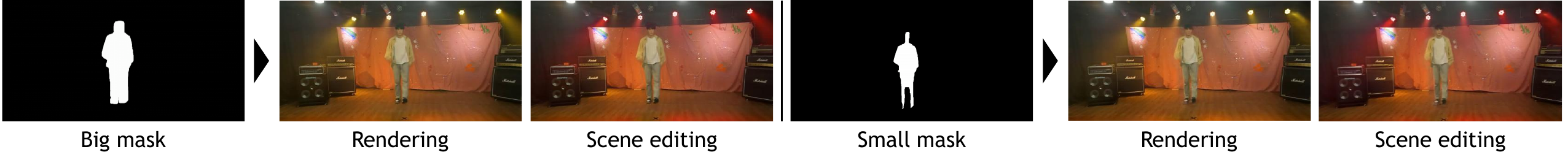}
     \caption{Results with various sizes of mask. The impact of the mask is marginal and our method robustly produce novel view and edited scene.}
     \label{fig:masking}
  \end{figure*}

\begin{figure}[t]
    \centering
     \includegraphics[width=\textwidth]{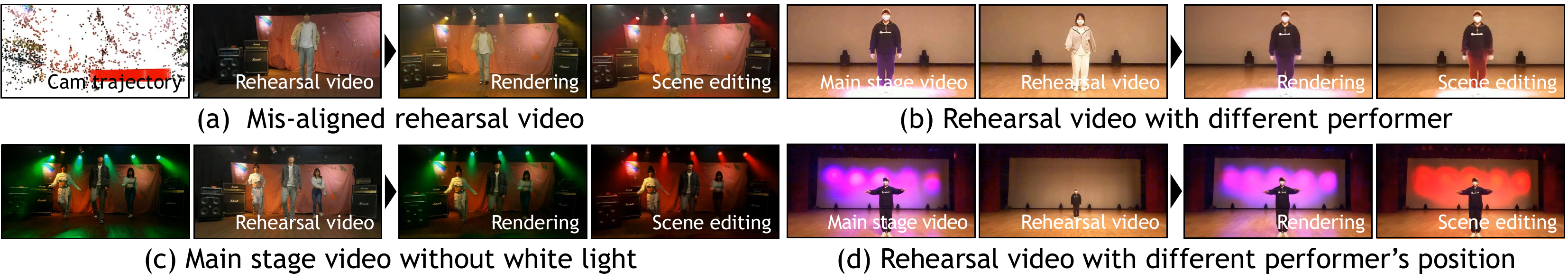}
     \caption{Results with various rehearsal prior conditions.}
     \label{fig:whitelight}
  \end{figure}

Due to several assumptions (eg, aligned cameras between rehearsal and main stages) in this pipeline, RehearsalNeRF may not seem to address various real-world scenarios. In this section, we address concerns about practicality and show the robustness of our RehearsalNeRF in many scenarios.

\noindent\textbf{Camera setups}\quad
Our work assumes that the cameras between the main stage and rehearsal stage are aligned, as it utilizes the difference in pixel values to initialize the illumination vector $v_{h}$. However, as demonstrated in~\Fref{fig:Cam_misalign}, even if this assumption is violated, the challenge can be easily addressed by synthesizing the Rehearsal videos for each camera position on the main stage through novel view synthesis.

\noindent\textbf{Complex lighting effects}\quad
In the hue initialization, the number of light sources determines the number of clusters in k-means clustering. This approach may seem ineffective in environments with a large number of light sources. However, the colors of the complex lighting can be represented by a few hue values.
To verify this, we carry out the real-world experiment in scenes in front of a large LED screen whose resolution $3,360 \times 1,620$~(5,443,200 light sources in total). As demonstrated in~\Fref{fig:LED}, RehearsalNeRF sufficiently represents these complex lighting effects by adjusting hyper-parameter $k$.

\noindent\textbf{Mis-sync between main and rehearsal video}\quad
The proposed real-world dataset consists of videos synchronized between the rehearsal stage and the main stage through the music. However, even when synchronization between the two stages is impossible, RehearsalNeRF still operates effectively due to the regularization for dynamic objects (\ref{sec:illumination}). To demonstrate this, we present the rendering outcomes for rehearsal videos shifted by +30 and +60 frames in~\Fref{fig:mis_sync} and~\Tref{tab:time_missync}. We also evaluate our RehearsalNeRF using a rehearsal prior that consists solely of the first frame image of the rehearsal video. It is noticeable that our RehearsalNeRF consistently produces high-quality results, regardless of aligned motion.

\noindent\textbf{Quality of mask}\quad
In \Fref{fig:masking}, we report qualitative results with different mask size. The size of masks are modulated by erosion and dilatation operation. Our method shows stable novel view synthesis and scene editing quality, even with largely varying mask sizes.

\noindent\textbf{Rehearsal prior conditions}\quad
To validate that our rehearsal prior does not require strict conditions, we conduct the following experiments, and they show plausible results in rendering and scene editing: 
(1) In \Fref{fig:whitelight} (a), we intentionally make a rehearsal video which is mis-aligned with the main stage video. To do this, we utilize 3D Gaussian splatting~\citep{kerbl20233d} to generate a video with an arbitrary pose;
(2) In \Fref{fig:whitelight} (b),
we use a rehearsal and main stage video whose performers are different. In addition, their clothes and motions are not identical. To handle this situation, we modify ~\Eref{eq: main rendering} by simply multiplying a learnable variable by $c_S$ and $c_D$ before rendering to account for their brightness;
(3) In \Fref{fig:whitelight} (c), 
we make the main stage video darker through a gamma correction. We simply multiply a learnable variable by $\mathbf{c}_S$ and $\mathbf{c}_D$ before rendering to account for their brightness. Since the white light changes the intensity of object's radiance regardless of their own colors, the learnable variable could compensate the loss of the white light in the main stage;
(3) In \Fref{fig:whitelight} (d), 
we show that our method works well in the extreme case where the performer's pose and distance from the camera array are significant changed between the main and rehearsal stages.

 \begin{figure*}[t]
    \centering
     \includegraphics[width=\textwidth]{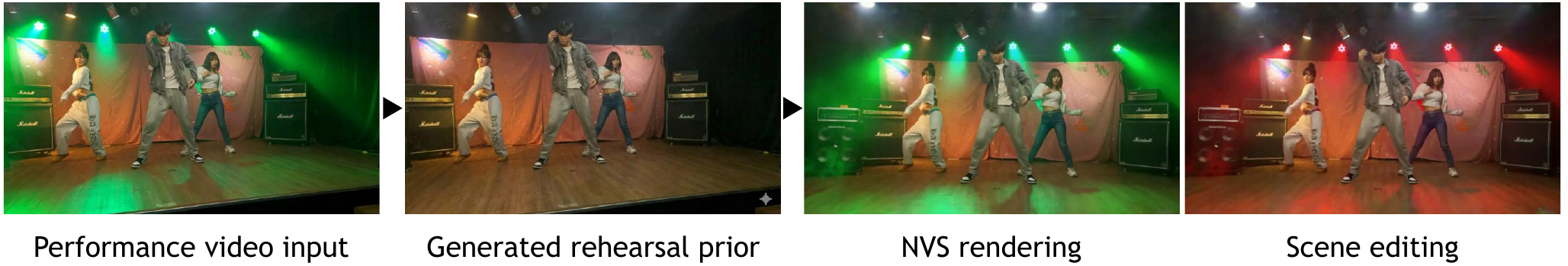}
     \caption{
     Qualitative results for the generative rehearsal prior. We demonstrate our rehearsal-free pipeline using a generative rehearsal prior. The rehearsal images are generated from a single frame of the input video without any lighting effects. Our model operates effectively with the generated rehearsal prior without performance degradation. While the rehearsal prior is essential to our model, its accuracy does not need to be strict, nor does it have to be captured from real data.
     }
     \label{fig:rehearsal_prior}
  \end{figure*}

\noindent\textbf{Generative rehearsal prior}\quad Although we propose and evaluate several strategies to relax this requirement, acquiring rehearsal data may still be challenging in practice. To address this issue more fundamentally, we further demonstrate that our algorithm can be applied using only the performance video by introducing a generative rehearsal prior. Specifically, we extract the first frame from the input video set (which contains highly dynamic lighting effects) remove its illumination using a generative model, and use the resulting images as a rehearsal prior. \Fref{fig:rehearsal_prior} presents both the generative rehearsal prior and the results produced by our rehearsal-video-free model. We use the Nano Banana Pro model served from Google Gemini-3 Pro. Despite slight residual illumination artifacts and multi-view inconsistencies in the generated rehearsal prior, our model produces plausible results without noticeable performance degradation. These results highlight the strong potential of the generative rehearsal prior for improving the generalization of our proposed method. Importantly, although the rehearsal prior plays a crucial role in our model, it does not require high-fidelity accuracy and can be obtained from generated data rather than real captures.

\section{Conclusion}
In this paper, we present a novel approach that trains and renders neural fields for dynamic scenes captured under drastically changing illuminations. 
The proposed method decouples dynamic lighting effects from static/moving objects. For this, our key idea is to devise the rehearsal prior, which can be easily taken before turning on lights for the main stage. 
In addition, we leverage the light color-adaptive vectors and the semantic-aware regulerizations to jointly optimize neural fields for static/dynamic objects and varying illuminations.
We demonstrate the effectiveness of the proposed method by showing impressive results on novel view synthesis and scene editing under dynamic illuminations. 

\noindent\textbf{Limitation}\quad
Directions for improvement exist. Since our work focuses on dynamic scene representations for short clips, efficient training schemes are needed for long sequence videos. 

\noindent\textbf{Data Availability}\quad
The authors declare that the data supporting the findings of this study are available in the supplementary information files. These files include the source code, video demos, and consent forms for using the performers’ portraits in the datasets. Furthermore, the real-world datasets reported in the manuscript are available from the first authors on reasonable request.

\section*{Acknowledgments}
This research was supported by CJ Corporation. This work was also supported by the Institute of Information \& Communications Technology Planning \& Evaluation (IITP) grants funded by the Korea government (MSIT) (RS-2020-II201361, No. RS-2024-00457882 and No.RS-2025-25441838, Development of a human foundation model for human-centric universal artificial intelligence and training of personnel) and the National Research Foundation of Korea(NRF) grant funded by the Korea government(MSIT)(RS-2024-00338439).

\section*{Declarations}
 Financial Interest: Changyeon Won, Hyunjun Jung and Seonmi Park received a Ph D stipend from Gwangju Institute of Science and Technology (GIST). Hae-Gon Jeon received salaries from GIST. Jungu Cho and Chi-Hoon Lee received salaries from CJ Corporation.

\begin{appendices}

\section{Application:  Confocal Fluorescence Microscopy}
\label{appendix:confocal}

\begin{table*}[h]
\centering
\caption{Similarity of input/output configurations between confocal fluorescence microscopy and RehearsalNeRF. }\label{tab:appendix_confocal_vs_ours}

\resizebox{\textwidth}{!}{
    \begin{tabular}{l|c c}
    \toprule
& \textbf{Confocal fluorescence microscope} & \textbf{RehearsalNeRF} \\
\midrule
\text{Illumination~~} & \text{various wavelength LEDs/Lasers} & \text{dynamic illuminations at a concert hall}\\
\text{Sensor} & \text{multiple detectors} & \text{multiple cameras}\\
\text{Subject} & \text{live cells and tissues} &  \text{performers}\\
\text{Output} & \text{~~3D volume rendering of live cells and tissues~~} & \text{~~radiance fields for components of a scene~~}\\
\bottomrule
\end{tabular}
}
\end{table*}

We demonstrate that our RehearsalNeRF can also be extended to other domains. As an example, we show a potential application to a confocal fluorescence microscope. 
Before describing the feasible scenario of our framework, we would like to introduce a concept of the confocal fluorescence microscope.
 
The confocal fluorescence microscope is an optical microscope to scan cells and tissues and to be used for studying the properties of them. As the name suggests, the microscopes use the fluorescence to generate an image with much higher intensity light sources which excite fluorescent species in a sample of interest. And, the confocal fluorescence microscope is devised to take higher resolutions and contrast images than that of the fluorescence microscope by blocking out-of-focus lights in image formulations.  Capturing multiple 2D images at either different depths~\citep{wang2021real} or various viewpoints (i.e. light field microscopy~\citep{browning2022microscopy}) in a sample enables 3D reconstruction and volume rendering of its structure. In particular, it becomes essential to investigate spatial arrangement of live cells and tissues with high precision in sciences, which is useful for assigning the localization to specific cellular compartments or finding out the relationships between them~\citep{jonkman2020tutorial}.

In total, we claim that the system configuration, inputs and outputs of the confocal fluorescence microscopes are very similar with those of our RehearsalNeRF as described in~\Tref{tab:appendix_confocal_vs_ours}.

\begin{figure*}[h]
\centering
\includegraphics[width=\textwidth]{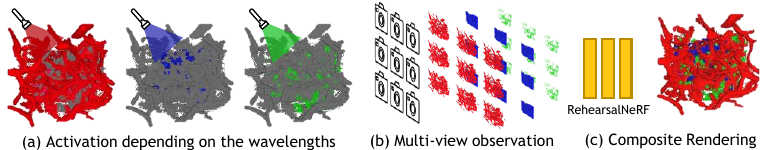}
\vspace{-4mm}
\caption{Applying RehearsalNeRF framework to confocal fluorescence microscopy. (a) Each sample type exhibits different activations according to the wavelengths of exposed lights. (b) Through multi-view observation such as light field microscopy, we can generate multi view images which can be used to train RehearsalNeRF. (c) We can render the 3D volume of the subject. Please check our supplementary video.}
\label{fig:microscopy}
\end{figure*}

 Following this scenario of the confocal fluorescence microscope, we provide an example to demonstrate the potential application. In~\cite{hontani2021multicolor}, a set of images for a mouse brain captured using a fluorescence microscopy is available. Using the images, we first make a composite image and then project it onto a regular grid for the multiview image acquisition, which becomes the input of our RehearsalNeRF. By training RehearsalNeRF with these images, we can synthesize novel views of specific cells or entire of cells.  As shown in~\Fref{fig:microscopy}, our RehearsalNeRF achieves the well-reconstructed spatial arrangement of the mouse brain. In the submitted video demo, we present rendering results that are synthesized by our RehearsalNeRF; please refer it. If this framework were extended to a temporal dataset consisting of flourescence microscope images, it could lead to the discovery of many interesting properties of cells and tissues by expanding into 4D domain (3D + time).

\section{Diversity of Our Dataset}

\begin{figure*}[ht]
\centering
\includegraphics[width=\textwidth]{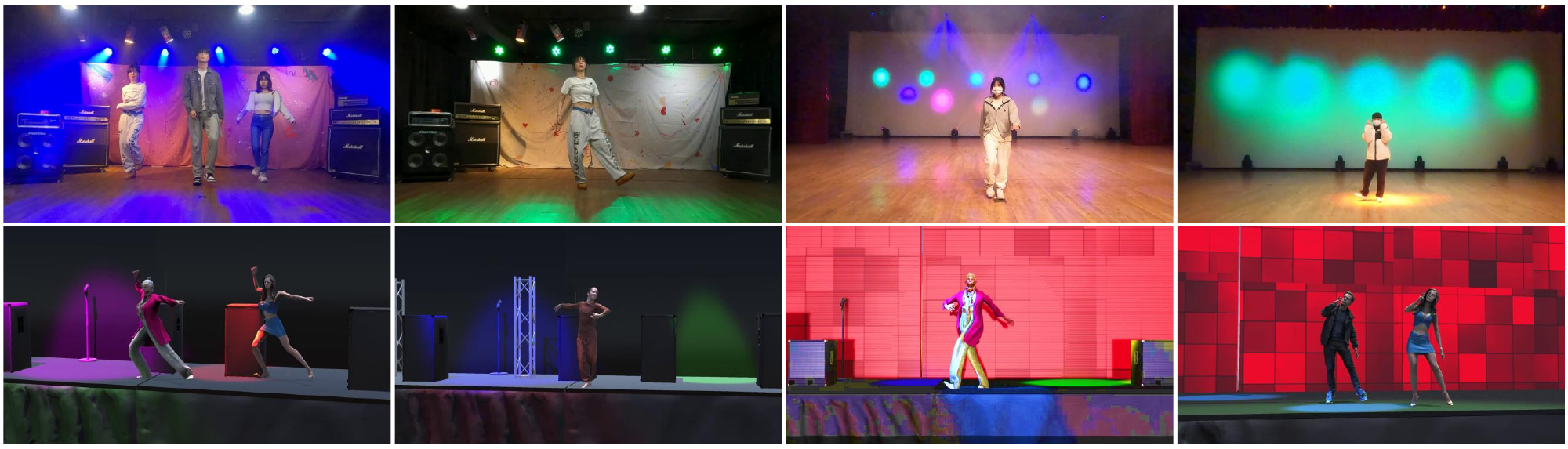}
\vspace{-4mm}
\caption{Examples of main stage video in our dataset. The real-world dataset are sampled in the upper row and the synthetic dataset in the lower row.}
\label{fig:dataset_eg_appendix}
\end{figure*}

 Since the advent of NeRF, new datasets have been taken whenever new problems are defined. We collect a new dataset with dynamic scenes with changing illumination, a unique situation that has not been addressed before. Our dataset ensures the diversity with respect to the number of dancers, choreography, costumes, and backgrounds, compared to the public datasets for NeRFs, as reported in~\Tref{tab:dataset_statistics} and ~\Fref{fig:dataset_eg_appendix}. As mentioned in the main manuscript, we also produce a synthetic dataset to test the robustness of our method for harder cases including intense light effects and highly saturated object surfaces. Furthermore, our dataset contains songs corresponding to the dances, which can be useful for multi-modal learning applications such as motion generation based on music.

\begin{table*}[h]
\centering
\small
\caption{Statistic of public datasets for NeRFs with dynamic motion and our dataset}
\resizebox{\textwidth}{!}{
\begin{tabular}{c||c|c|c}
\hline
Dataset                                       & \# of scenes             & \# of dynamic subjects   & \# of backgrounds        \\
\hline
{Nerfies Validation Rig~\citep{park2021nerfies}}
& {4} & {3} & {3} \\
{Neural 3D video~\citep{li2022neural}}
& {6} & {1} & {2} \\
{Meet Room~\citep{li2022streaming}}
& {3} & {3} & {1} \\

\hline
{Ours (real-world)}
& {10} & {4} & {3} \\
{Ours (synthetic)}
& {4} & {4} & {4} \\
\hline

\end{tabular}  
}
\label{tab:dataset_statistics}
\end{table*}
\section{More Experimental Results}
\subsection{Video manipulation}

\begin{figure*}[t]
\centering
\includegraphics[width=\textwidth]{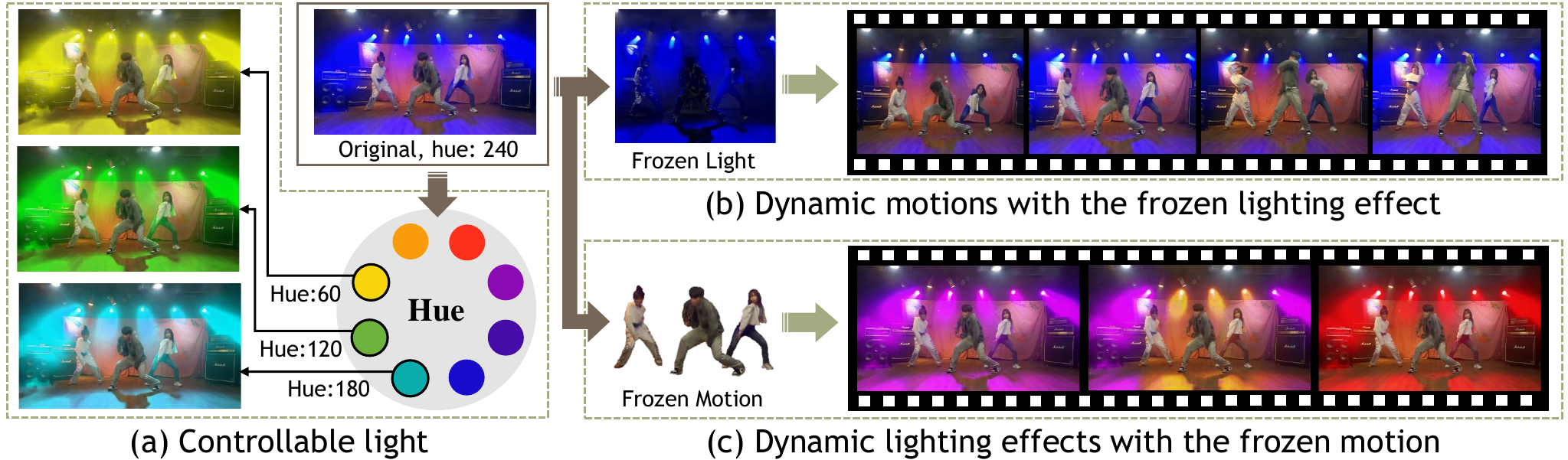}
\caption{Applications of RehearsalNeRF to video manipulation. (a) We can change the light colors after recording the video. (b) Our RehearsalNeRF produces the natural-looking dance video under artificially stopping the light effects. (c) We show a video with the frozen motion, but the light sources still work. Please check our supplementary video.}
\label{fig:application}
\end{figure*}

Accurate disentanglement of neural fields for dynamic illuminations and scene objects in the time domain can facilitate many applications.
As examples, we show video editing applications such as time stop effect (a.k.a bullet-time) by manipulating the temporal flow of the decoupled neural fields.

A controllable light effect that tunes light colors after taking the video is one of the applications of our RehearsalNeRF. Thanks to the accurate neural representation for dynamic lights, we can artificially direct the various light color effects in \Fref{fig:application} (a). Another interesting application is to freeze the subject's motion and light colors. As shown in \Fref{fig:application} (b) and (c), since the illumination neural fields are supervised independently for the motion and geometry of the subject, we can easily produce the videos that show natural-looking motions in artificially fixed light conditions and vice versa. 

\subsection{Analysis on synthetic dataset}

\begin{figure*}[h]
\centering
\includegraphics[width=\textwidth]{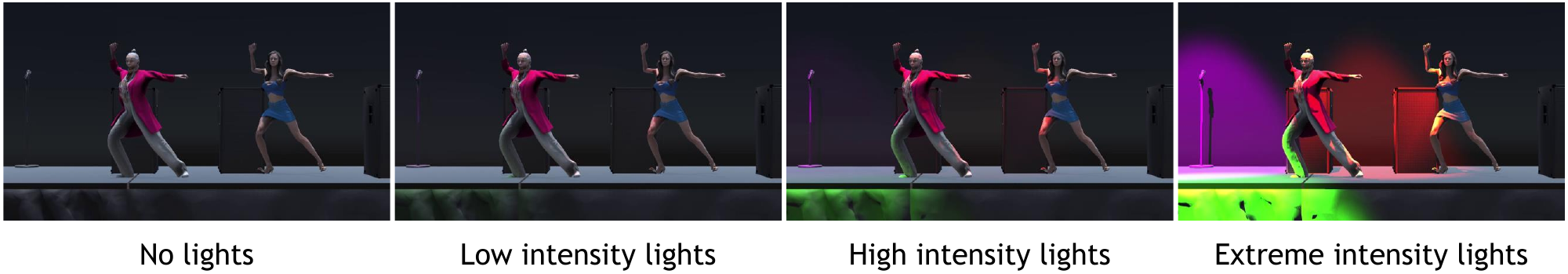}
\vspace{-4mm}
\caption{Examples of synthetic datasets under various light intensities in dynamic illumination. From left to right, we incrementally increase the intensity of lights. The intensity predominantly used in our main paper corresponds to the 'high intensity' setting, as shown in the third column.}
\label{fig:supp_synthetic_intensity}
\end{figure*}

The synthetic dataset is designed to simulate high-intensity lighting setups, which can lead to texture degradation and color saturation in objects. This enables us to test our RehearsalNeRF in scenarios where real-world datasets cannot cover. In our experiments, we evaluate the performance of our model and compare it with the comparison methods under varying intensities of dynamic illumination. As depicted in \Fref{fig:supp_synthetic_intensity}, we have created a synthetic dataset from no lighting to the extreme lighting conditions. We then assess the PSNR and SSIM values for novel view synthesis on the scene illustrated in \Fref{fig:supp_synthetic_intensity}.

\begin{figure*}[h]
\centering
\includegraphics[width=\textwidth]{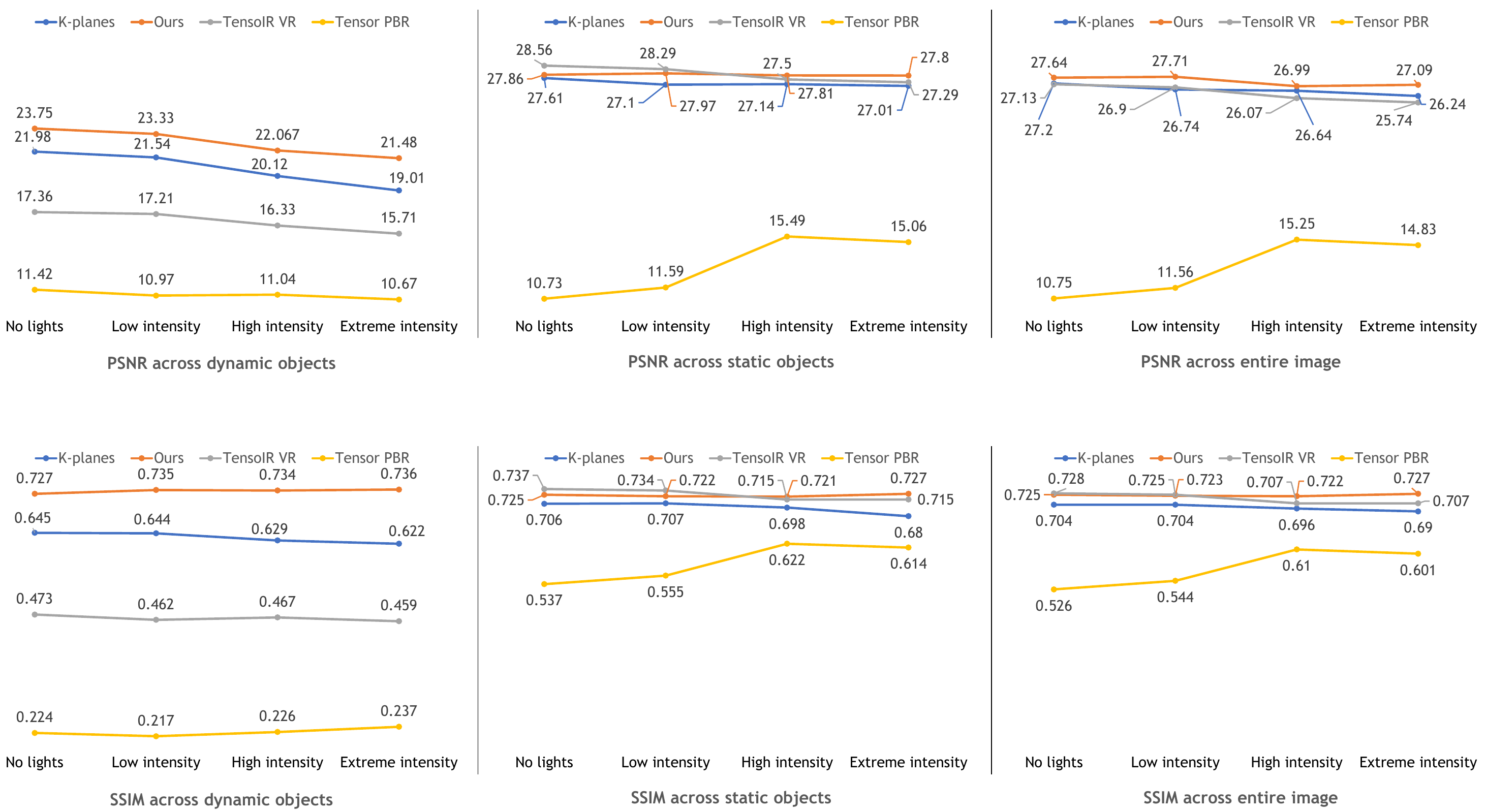}
\vspace{-4mm}
\caption{Quantitative results for the synthetic dataset under the lighting conditions with varying intensities. We evaluate the results across dynamic objects, static objects, and the entire image.}
\label{fig:supp_graph}
\end{figure*}

In Figure \ref{fig:supp_graph}, we present the results of novel view synthesis for only dynamic objects, only static objects and the entire image. We employ ground truth masks for this evaluation, and compare ours with K-planes~\citep{fridovich2023k} and TensorIR~\citep{jin2023tensoir}. Notably, the significant performance gap between our RehearsalNeRF and the other methods is observed when assessments are confined to dynamic objects. This implies that dynamic illumination significantly hinders learning dynamic objects in neural fields. Moreover, the reason why our RehearsalNeRF exhibits the best performance on dynamic objects is that the $\mathcal{L}_{dyn}$ works well for the shapes of dynamic objects. This capability stems from the fact that the loss term leverages the rehearsal priors and optical flow, which provide coarse clues for shape reconstruction.
The performance gap becomes noticeable as the intensity increases. This trend is consistent in the dynamic objects as well as both the static objects and the entire image. In conclusion, our RehearsalNeRF is robust to handling intense lighting conditions as well.

\section{Computational resource}

\begin{table*}[t]
    \centering
    \caption{Comparison of computational efficiency. We report the peak GPU memory usage during inference, training time, and inference time measured on a single RTX 3090 GPU.}
    \label{tab:efficiency_comparison}
    \resizebox{0.8\linewidth}{!}{
        \begin{tabular}{lccc}
            \toprule
            \textbf{Method} & \textbf{VRAM (MB)} & \textbf{Training Time (sec)} & \textbf{Inference Time (sec)} \\
            \midrule
            D$^2$NeRF       & 3,739  & 16,406 & 72.10 \\
            K-Planes        & 1,081  & 10,491 & 7.93 \\
            TensoIR VR      & 2,449  & 27,110 & 3.54 \\
            TensoIR PBR     & 15,641 & 27,110 & 66.70 \\
            \midrule
            \textbf{Ours}   & 1,639  & 32,250 & 14.80 \\
            \bottomrule
        \end{tabular}
    }
\end{table*}

We report runtime and resource analysis as your request in \Tref{tab:efficiency_comparison}. For a fair comparison, we standardize the batch size to 1024 rays and conduct all experiments on the synthetic dataset using a single NVIDIA RTX 3090 GPU. We report the peak GPU memory usage as measured by \texttt{nvidia-smi}. Our model exhibits a longer training time compared to other methods due to its architectural complexity. Addressing this limitation is left for future work. Possible directions include incorporating geometric priors or adopting more fast coordinate-based methods, such as Instant-NGP.

\end{appendices}

\bibliography{sn-bibliography}

\end{document}